\documentclass[11]{article}
\usepackage{fullpage}
\usepackage[utf8]{inputenc} 
\usepackage[T1]{fontenc}    
\usepackage{url}            
\usepackage{booktabs}       
\usepackage{amsfonts}       
\usepackage{nicefrac}       
\usepackage{microtype}      

\usepackage{algorithm}
\usepackage{algorithmic}
\usepackage{graphicx}
\usepackage{subfigure}
\usepackage{amsmath,amssymb,amsthm,graphicx}
\usepackage{epsfig}
\usepackage{psfrag}
\usepackage{enumerate} 
\usepackage{setspace}
\usepackage{float} 
\usepackage{color}
\usepackage{array}
\usepackage{pgf,tikz}
\usepackage{mathrsfs}
\usepackage{multirow}
\usepackage{wrapfig}
\usepackage{makecell}

\usetikzlibrary{arrows}
\definecolor{ffqqqq}{rgb}{1.,0.,0.}
\definecolor{xfqqff}{rgb}{0.4980392156862745,0.,1.}

\usepackage{gensymb}
\usepackage{diagbox}
\usepackage{xspace}
\usepackage{arydshln}
\usepackage{enumitem}
\usepackage{wrapfig}

\newcommand{\searchset}[1]{S^{#1}}

\newcommand{\px}{\textrm{px}}

\newcommand{\EE}{\mathbb{E}\:}

\newcommand{\Pert}{\ensuremath{\Delta}}

\newcommand{\Pertset}{\mathcal{S}}
\newcommand{\Trafo}[2]{\mathcal{T}(#1, #2)}

\newcommand{\YID}{\ensuremath{Y}}
\renewcommand{\P}{\mathbb{P}}
\newcommand{\loss}{\ell}
\newcommand{\func}{f}

\newcommand{\Loss}[1]{\mathcal{L}_{#1}}
\newcommand{\PopLoss}[1]{\Loss{#1}}
\newcommand{\PopLossReg}[1]{\PopLoss{#1}}

\DeclareMathOperator*{\argmax}{arg\,max}
\DeclareMathOperator*{\argmin}{arg\,min}
\newcommand{\RN}{\mathbb{R}}

\newcommand{\X}{\mathcal{X}}

\newcommand{\std}{\ensuremath{\text{std}}}
\newcommand{\stdaug}{\ensuremath{\std^\star}}
\newcommand{\rnd}{\text{rnd}}

\newcommand{\at}{\text{Adversarial training}}

\newcommand{\wokabb}{\text{Wo-$\numwok$}}
\newcommand{\woabb}[1]{\text{Wo-${#1}$}}
\newcommand{\wok}{\text{worst-of-$\numwok$}}
\newcommand{\wo}[1]{\text{worst-of-${#1}$}}

\newcommand{\numwok}{k}
\newcommand{\fo}{\text{S-PGD}}

\newcommand{\fstar}{f^\star}
\newcommand{\suppmat}{Appendix}

\newcommand{\defvar}{\texttt{def}}

\newcommand{\eps}{\epsilon}

\newcommand{\Group}{\mathbb{G}}
\newcommand{\defn}{:=}
\newcommand{\Gorbit}[1]{\Group(#1)}
\newcommand{\SubGorbit}[1]{G^{#1}}
\newcommand{\SubGroup}[1]{\Group^{#1}}

\newcommand{\ortho}{\perp \!\!\! \perp }
\newcommand{\InvF}{\mathcal{V}}
\newcommand{\Fspace}{\mathcal{F}}

\newcommand{\nat}{\text{nat}}
\newcommand{\mix}{\text{mix}}

\newcommand{\Reg}{\ensuremath{R}}

\newcommand{\frob}{f^{\text{rob}}}
\newcommand{\abest}{a^\star}
\newcommand{\Pertdiam}{\Pert^{\star}}
\newcommand{\regfunc}{h}
\newcommand{\regadvnat}{\tilde{R}}
\newcommand{\AT}{\text{AT}}

\newcommand{\ALP}{\text{ALP}}
\newcommand{\tr}{\text{KL}}
\newcommand{\trcce}{\text{KL-C}}
\newcommand{\ltwo}{\ensuremath{\ell_2}}
\newcommand{\alp}{ALP}

\newcommand{\KL}{\text{D}_{\text{KL}}}

\newcommand{\stn}{\text{STN}}
\newcommand{\grn}{\text{GRN}}
\newcommand{\fnat}{\ensuremath{f^{\text{nat}}}}

\newcommand{\batchtype}{\texttt{batch}}

\newcommand{\regtype}{\texttt{Reg}\:}

\newcommand{\gresnet}{\text{G-ResNet44}}
\newcommand{\etn}{\text{ETN}}
\newcommand{\aug}[1]{\ensuremath{{#1}^{\star}}}
\newcommand{\twostage}{\text{STN+}}

\newcommand{\Gspace}{\mathcal{I}}
\newcommand{\SetOrbit}{\mathcal{G}}
\newcommand{\orbitfunc}{\Psi}

\newcommand{\Frob}{F^{\rob}}
\newcommand{\rob}{\text{rob}}
\newcommand{\fmin}{f^{\min}}

\newcommand{\Xsupp}{\widetilde{\X}}
\theoremstyle{plain}

\newtheorem{theo}{Theorem}

\theoremstyle{definition}

\usepackage{hyperref}       

\begin{document}

\begin{center}
{\bf{\Large{Invariance-inducing regularization using worst-case transformations suffices to boost accuracy and spatial robustness}}}

\vspace{.2in} 

{\large{
    \begin{tabular}{ccccc}
      Fanny Yang$^{1,3}$ &&  Zuowen Wang && Christina Heinze-Deml$^{2}$
\end{tabular}
}}

\vspace{.1in} 

\begin{tabular}{ccc}
  Institute for Theoretical Studies$^1$ && Department of Statistics$^3$\\
  Seminar for Statistics$^2$  && Stanford University \\
  ETH Zurich, Switzerland  && 
\end{tabular}

\vskip 0.3in

\begin{abstract}

  This work provides theoretical and empirical evidence that
  invariance-inducing regularizers can increase predictive accuracy
  for worst-case spatial transformations (spatial \emph{robustness}).
  Evaluated on these \emph{adversarially} transformed examples, we
  demonstrate that adding regularization on top of standard or
  adversarial training reduces the relative error by 20\% for CIFAR10
  without increasing the computational cost.  This outperforms
  handcrafted networks that were explicitly designed to be
  spatial-equivariant. Furthermore, we observe for SVHN, known to have
  inherent variance in orientation, that robust training also improves
  standard accuracy on the test set. We prove that this no-trade-off
  phenomenon holds for adversarial examples from \emph{transformation}
  groups in the infinite data limit.
  
 \end{abstract}
\end{center}

\section{Introduction}
\label{sec:intro}

As deployment of machine learning systems in the real world has
steadily increased over recent years, the trustworthiness of these systems is of crucial importance.  This is particularly the case for safety-critical
applications. For example, the vision system in a self-driving car should correctly
classify an obstacle or human irrespective of their orientation.
Besides being relevant from a security perspective,
a measure for spatial invariance also helps to
gauge interpretability and reliability of a model. If an image of
a child rotated by $10^{\circ}$ is classified as a trash can, can we
really trust the system in the wild?


As neural networks have been shown to be expressive both theoretically
\cite{Hornik89,Barron93,Hanin17} and empirically \cite{Zhang16}, in
this work we study to what extent standard neural networks predictors
can be made invariant to small rotations and translations. In contrast
to enforcing conventional invariance on entire group orbits, we weaken
the goal to invariance on smaller so-called \emph{transformation
  sets}.
This requirement reflects the aim to be invariant to transformations
that do not affect the labeling by a human.  During test time we
assess transformation set invariance by computing the prediction
accuracy on the worst-case (\emph{adversarial}) transformation in the
(small) transformation set of each image in the test data.
The higher this worst-case prediction accuracy of a model is, the more
\emph{spatially robust} we say it is. 
Importantly, we use the same
terminology as in the very active field of adversarially robust
learning \cite{Szegedy13, Moosavi16, Kurakin16,
  Papernot17,Carlini17,Madry18, Samangouei18, Sinha18,
  Ragunathan18,Wong18,Mirman18}, but we consider adversarial examples
with respect to \emph{spatial} instead of $\ell_p$-transformations of
an image.



Recently, it was observed in \cite{Engstrom17,Fawzi15, Pei17,Kanbak18, Geirhos18, Alcorn18} that worst-case prediction performance 
drops dramatically for neural network classifiers obtained
using standard training, even for rather small transformation sets.
In this context, we examine the effectiveness of regularization that
explicitly encourages the predictor to be constant for transformed
versions of the same image, which we refer to as being \emph{invariant} on
the transformation sets.  Broadly speaking, there are two approaches
to encourage invariance of neural network predictors. On the one hand,
the relative simplicity of the mathematical model for rotations and
translations has led to carefully hand-engineered architectures that
incorporate spatial invariance directly \cite{Jaderberg15,
  Pollefeys16, Cohen16, Marcos17, Worrall17, Weiler18, Esteves18,
  Tai19}.
On the other hand, augmentation-based methods \cite{Baird92, Yaeger97}
constitute an alternative approach to encourage desired invariances on
transformation sets.  Specifically, the idea is to augment the
training data by a random or smartly chosen transformation of every
image for which the predictor output is enforced to be close to the
output of the original image. This \emph{invariance-inducing}
regularization term is then added to the cross entropy loss 
for back-propagation.


While augmentation-based methods can be used out of the box whenever
it is possible to generate samples in the transformation set of
interest, it is unclear how they compare to architectures that are
tuned for the \emph{particular} type of transformation using prior
knowledge. Studying robustness against spatial transformations in
particular allows us to compare the robust performance of these two
approaches, as spatial-equivariant networks have been somewhat
successful in enforcing invariance. In contrast, this cannot be claimed for
higher-dimensional $\ell_p$-type perturbations.
In the empirical sections of
this paper, we hence want to explore the following questions:
\begin{enumerate}
\item To what extent can augmentation and regularization based methods improve spatial robustness of common deep neural networks?
\item How does augmentation-based invariance-inducing regularization perform in case of small spatial transformations compared to representative specialized architectures designed to achieve invariance against entire transformation groups?
\end{enumerate}

As a justification for employing this form of \emph{invariance-inducing}
regularization, we prove in our theoretical section~\ref{sec:theory} that when
perturbations come from transformation groups, predictors that
optimize the robust loss are in fact invariant on the set of
transformed images.  Although recent works show a fundamental
trade-off between robust and standard accuracy in constructed $\ell_p$
perturbation settings \cite{Tsipras18, Zhang19, RXYDL19}, we
additionally show that this is fundamentally different for spatial
transformations due to their group structure.

For the empirical study, we implemented various
augmentation based training methods as described in
Sec.~\ref{sec:experimental_setup}.  In Sec.~\ref{sec:results}, we
compare spatial robustness for augmentation-based methods and
specialized neural network architectures on CIFAR-10 and SVHN.
Although group-invariance should automatically imply robust
predictions against \textit{all} transformations in the group,
group-equivariant networks have not been extensively evaluated using
adversarially chosen, but rather random transformations.
In experiments with CIFAR-10 and SVHN, we find that regularized
methods can achieve $\sim20\%$ relative adversarial error reduction
compared to previously proposed augmentation-based methods (including
adversarial training) without requiring additional computational
resources. Furthermore, they even
outperform representative handcrated networks that were
explicitly designed for invariance.

\section{Theoretical results for invariance-inducing regularization}
\label{sec:theory}

In this section, we first introduce our notion of transformation sets
and formalize robustness against a small range of translations and
rotations.  We then prove that, on a population level, constraining or
regularizing for transformation set invariance yields models that
minimize the robust loss.  Moreover, when the label distribution is
constant on each transformation set, we show that the set of robust
minimizers not only minimizes the natural loss but, under mild
conditions on the distribution over the transformations, is even
equivalent to the set of natural minimizers.

Although the framework can be applied to general problems and
transformation groups, we consider image classification for
concreteness.  In the following, $X \in \X \subset \RN^d$ are the observed images,
 $Y\in\RN^p$ is the one-hot vector for multiclass labels and both are
random variables from a joint distribution $\P$. The function $f:\RN^d
\to \RN^p$ in function space $\Fspace$ (e.g. deep neural network in
experiments) maps the input image to a logit vector that is then
used for prediction via a softmax layer.

\subsection{Transformation sets}
Invariance with respect to spatial transformations is often thought of in terms of group
equivariance of the representation and prediction.  Instead of
invariance with respect to all spatial transformations in a group, we
impose a weaker requirement, that is invariance against transformation
sets, defined as follows.  We denote by $\SubGorbit{z}$ a compact
subset of images in the support of $\P$ that can be obtained by
transformation of an image $z \in \X$.  $\SubGorbit{z}$ is called a
\emph{transformation set}. For example in the case of rotations, the
transformation set $\SubGorbit{z}$ corresponds to the set of observed
images in a dataset that are different versions of the same image $z$,
that can be obtained by small rotations of one another.

By the technical assumption on the space
of real images that the sampling operator is bijective, the mapping $z
\to \SubGorbit{z}$ is bijective.  We can hence define
$\SetOrbit$, a set of transformation sets, by $\SetOrbit =
\cup_{z\in\X} \SubGorbit{z}$ for a given transformation group. Importantly,
the bijectivity assumption also leads to $\SubGorbit{z}$ being
disjoint for different 
images $z\in \X$. The above definition is distribution dependent and
$\SetOrbit$ partitions the support $\Xsupp$ of the distribution.
More details on the aforementioned concepts and
definitions can be found in Sec.~\ref{sec:groups} in the \suppmat.


We say that a function $f$ is \emph{(transformation-)invariant} if
$f(x) = f(x')$ for all $x,x'\in U$ for all $U \in \SetOrbit$ and
denote the class of all such functions by $\InvF$.
Using this notation, fitting a model with high accuracy under worst-case
``small'' transformations of the input can be mathematically captured
by the robust optimization formulation~\cite{BenTal09} of minimizing
the \emph{robust loss}
\begin{equation}
  \label{eq:advmin}
  \PopLoss{\rob}(f) :=  \EE_{X,Y} \sup_{x' \in\SubGorbit{X}} \loss(f(x'), Y)
\end{equation}
in some function space $\Fspace$.
We call the solution of this problem the (spatially) \emph{robust} minimizer.
While adversarial training aims to optimize the empirical version of
Eq.~\eqref{eq:advmin}, the converged predictor might be far from the
global population minimum, in particular in the case of nonconvex optimization 
landscapes encountered when training neural networks. Furthermore, we show in the following
section that for robustness over transformation sets, constraining
the model class to invariant functions leads to the same optimizer of
the robust loss.
These facts motivate invariance-inducing regularization which we then show to
exhibit improved robust test accuracy in practice.

\subsection{Regularization to encourage invariance}

For any regularizer $\Reg$, we define the corresponding constrained set of functions $\InvF(\Reg)$ as
\begin{equation*}
  \InvF(\Reg) := \{f: \Reg(f, x, y) = 0 \quad \forall (x,y) \in \text{supp}(\P)\},
\end{equation*}
where $\text{supp}(\P)$ denotes the support of $\P$.  When $\Reg(f, x,
y) = \sup_{x'\in \SubGorbit{x}} \regfunc(f(x), f'(x))$ and $\regfunc$
is a semimetric\footnote{The weaker notion of a semimetric satisfies
  almost all conditions for a metric without having to satisfy the
  triangle inequality.} on $\RN^p$, we have $\InvF(\Reg) = \InvF$.
We now consider constrained optimization problems of the form
\begin{align}
  &\min_{f\in\Fspace} \EE \loss(f(X), Y) \text{ s.t. }  f\in \InvF(\Reg), \tag{O1}  \label{eq:regnat} \\
  &\min_{f\in\Fspace} \EE  \sup_{x' \in\SubGorbit{X}} \loss(f(x'), Y) \text{ s.t. }  f\in \InvF(\Reg) \tag{O2} \label{eq:regadv}.
\end{align}
The following theorem shows that~\eqref{eq:regnat}, \eqref{eq:regadv}
are equivalent to~\eqref{eq:advmin} if the set of all invariant
functions $\InvF$ is a subset of the function space $\Fspace$.
\begin{theo}
  \label{theo:1}
  If $\InvF \subseteq \Fspace$, 
  all minimizers of the adversarial
  loss~\eqref{eq:advmin} are in $\InvF$.  If
  furthermore $\InvF(\Reg) \subseteq \InvF$, any solution of the
  optimization problems~\eqref{eq:regnat}, \eqref{eq:regadv} minimizes the adversarial loss.
\end{theo}
The proof of Theorem~\ref{theo:1} can be found in the \suppmat\,in
Sec.~\ref{sec:theo1proof}. 
Since exact projection onto the constrained set is in general not
achievable for neural networks,
an alternative method to induce invariance is to relax the constraints
by only requiring $f\in \{f: \Reg(f, x, y) \leq \eps \quad \forall (x,
y) \in \text{supp}(\P)\}$. Using Lagrangian duality, \eqref{eq:regnat}
and \eqref{eq:regadv} can then be rewritten in penalized form for some
scalar $\lambda > 0$ as
\begin{align}
  &\min_{f\in\Fspace} \PopLossReg{nat}(f; \Reg, \lambda) := \min_{f\in\Fspace} \EE \loss(f(X),Y) + \lambda\Reg(f, X, Y), \label{eq:natpen} \\
  &\min_{f\in\Fspace} \PopLossReg{rob}(f; \Reg, \lambda) := \min_{f\in\Fspace} \EE \sup_{x'\in\SubGorbit{X}}\loss(f(x'),Y) + \lambda\Reg(f, X, Y).    \label{eq:advpen}
\end{align}

In Sec.~\ref{sec:diffreg} we discuss how ordinary adversarial
training, and modified variants that have been proposed thereafter, can
be viewed as special cases of Eqs.~\eqref{eq:natpen} and~\eqref{eq:advpen}. 
On the other hand, the constrained regularization formulation
corresponds to restricting the function space and is hence comparable
with hand-crafted network architecture design as described in
Sec.~\ref{sec:spatnet}. 



\subsection{Trade-off between natural and robust accuracy}

Even though high robust accuracy~\eqref{eq:advmin} might be the main
goal in some applications, one might wonder whether the robust
minimizer exhibits lower accuracy on untransformed images
(\emph{natural accuracy}) defined as $\PopLoss{\nat}(f) := \EE_{X,Y}
\loss(f(X),Y)$ \cite{Tsipras18, Zhang19}. In this section we address this question and identify the conditions for transformation set
perturbations under which minimizing the robust loss does not lead to
decreased natural accuracy. Notably, it even increases under mild assumptions.

One reason why adversarial examples have attracted a lot of interest
is because the prediction of a given classifier can change in a
perturbation set in which all images appear the same to the human eye.
Mathematically, in the case of transformation sets, the latter can be
modeled by a property of the true distribution. Namely, it translates into the
conditional distribution $Y$ given $x$, denoted by $\P_{\SubGorbit{x}}$, being constant for all $x$
belonging to the same subset $U\in\SetOrbit$. In other words, $Y$ is conditionally
independent of $X$ given $\SubGorbit{X}$, i.e.\ $Y\ortho
X|\SubGorbit{X}$. Under this assumption the next theorem shows that
there is no trade-off in natural accuracy for the transformation
robust minimizer.
\begin{theo}[Trade-off natural vs. robust accuracy]
  \label{theo:tradeoff}
  Under the assumption of Theorem~\ref{theo:1} and if $Y \ortho X |
  \SubGorbit{X}$ holds, the adversarial minimizer also minimizes the
  natural loss. If moreover, $\P_{\SubGorbit{z}}$ has support
  $\SubGorbit{z}$ for every $z\in \Xsupp$ and the loss $\loss$ is injective, then every
  minimizer of the natural loss also has to be invariant.
\end{theo}

As a consequence, minimizing the constrained optimization problem
\eqref{eq:regnat} could potentially help in finding the optimal
solution to minimize standard test error.  Practically, the
assumption on the distribution of the transformation sets
$\SubGorbit{z}$ corresponds to assuming non-zero inherent transformation variance in the
natural distribution of the dataset.
In practice, we indeed observe a boost in natural accuracy for robust
invariance-inducing methods in Sec.~\ref{sec:results} on SVHN, a
commonly used benchmark dataset for spatial-equivariant networks for
this reason.

One might wonder how this result relates to several recent publications such
as \cite{Tsipras18, Zhang19} that presented toy examples for
which the $\ell_\infty$ robust solution must have higher natural loss
than the Bayes optimal solution even in the infinite data limit.
On a fundamental level, $\ell_\infty$ perturbation sets are of different
nature compared to transformation sets on generic distributions of $\X$. 
In the distribution considered in \cite{Tsipras18, Zhang19}, there is no unique mapping
from $x\in \X$ to a perturbation set and thus the conditional independence
property does not hold in general. 

\subsection{Different regularizers and practical implementation}
\label{sec:diffreg}
 In order to improve robustness against spatial transformations we
 consider different choices of $\Reg(f,x,y)$ in the regularized
 objectives~\eqref{eq:natpen} and~\eqref{eq:advpen} that we then compare empirically in
 Sec.~\ref{sec:results}.
 This allows us to view
 a number of variants of adversarial training in a unified
 framework. Broadly speaking, each approach listed below consists of
 first searching an adversarial example according to some mechanism
 which is then included in a \emph{regularizing function}, often some
 weak notion of distance between the prediction at $X$ and the new example.
The following choices of regularizers involve the maximization of a
regularizing function over the transformation set
\begin{align*}
  \Reg_{\AT}(f, X, Y) &= \sup_{x'\in\SubGorbit{X}} \loss(f(x'),Y) -\loss(f(X),Y) \text{ (equivalent to \cite{Szegedy13,Madry18} for $\PopLoss{\nat}$)} \\
  \Reg_{\ltwo}(f, X, Y) &= \sup_{x'\in\SubGorbit{X}} \|f(X)-f(x')\|^2_2 \\ 
  \Reg_{\tr}(f, X, Y) &= \sup_{x'\in\SubGorbit{X}} \KL(f(x'), f(X)) \text{ (equivalent to \cite{Zhang19} for $\PopLoss{\nat}$)\footnote{Notice that \cite{Zhang19} is actually using the KL divergence in their implementation, which is not the same as cross-entropy as claimed in the paper.}}
  \end{align*}
where $\KL$ is the KL divergence on the softmax of the (logit) vectors
$f \in \RN^p$. In all cases we refer to the maximizer as an
\emph{adversarial example} that is found using \emph{defense
  mechanisms} as discussed in Section~\ref{sec:attacks_defenses}. Note
that for $\Reg_{\ltwo}$ and $\Reg_{\tr}$ the assumption $\InvF(\Reg)
\subseteq \InvF$ in Theorem~\ref{theo:1} is satisfied.

Instead of performing a maximization of the regularizing function to
find the adversarial example $x'$, we can also choose $x'$ in
alternative ways
The following variants are
explored in the paper, two of which are reminiscent of previous work
\begin{align*}
  \Reg_{\ALP}(f, X, Y) &= \|f(x')-f(X)\|_2^2 \:\: \text{ with } \:\: x' = \argmax_{u\in\SubGorbit{X}} \loss(f(u),Y) \text{  (equivalent to \cite{Kannan18})}\\
   \Reg_{\trcce}(f, X, Y) &= \KL(f(x'), f(X)) \:\: \text{ with }\:\: x' = \argmax_{u\in\SubGorbit{X}} \loss(f(u),Y)\\
  \Reg_{\regfunc-DA}(f, X) &= \EE_{x'\in\SubGorbit{X}} \regfunc(f,X,X') \text{ (similar to \cite{Heinze17})}
\end{align*}
The last regularizer suggests using an additive
penalty on top of data augmentation, with either one or even multiple
random draws, where the penalty can be any of the above semimetrics
$\regfunc$ between $f(X)$ and $f(x')$, such as the $\ell_2$ or $\KL$ distance.
Albeit suboptimal, the experimental
results in Section~\ref{sec:results} suggest that simply adding the
additive regularization penalty on top of randomly drawn data matches
general adversarial training in terms of robust prediction at a
fraction of the computational cost.
In addition, Theorem~\ref{theo:tradeoff} suggests that even
when the goal is to \emph{improve standard accuracy} and one expects
inherent variance of nuisance factors in the data distribution
it is likely helpful to use regularized data augmentation with $\Reg_{\regfunc-DA}$
instead of vanilla data augmentation. Empirically we observe this
on the SVHN dataset in Section~\ref{sec:results}.



{\bf Adversarial example for spatial transformation sets} Since
$\SubGorbit{X}$ is not a closed group and we do not even know whether
the observation $X$ lies at the boundary of $\SubGorbit{X}$ or in the
interior, we cannot solve the maximization constrained to
$\SubGorbit{X}$ in practice. However, for an appropriate choice of set
$\Pertset$, we can instead
minimize an upper bound of \eqref{eq:advmin} which reads
\begin{equation}
  \label{eq:advminapp}
  \min_{f\in\Fspace} \EE \sup_{\Pert \in \Pertset} \loss(f(\Trafo{X}{\Pert}), Y) \geq \min_{f\in\Fspace} \EE \sup_{x' \in\SubGorbit{X}} \loss(f(x'), Y)
\end{equation}
where $\Pertset$ is the set of transformations that we search over and
$\Trafo{X}{\Pert}$ denotes the transformed image with transformation
$\Pert$ (see Sec.~\ref{sec:groups} in the \suppmat\ for an explicit
construction of the transformation search set $\Pertset$). The left hand side
in~\eqref{eq:advminapp} is hence what we aim to solve in practice
where the expectation is over the empirical joint distribution of
$X,Y$. The relaxation of $\SubGorbit{X}$ to a range of transformations
of $X$ that is $\{ \Trafo{X}{ \Pert }\ : \Pert \in \Pertset \}$ is
also used for the maximization within the regularizers.

In Figure~\ref{fig:ex_trafo} one pair of example images is shown: the original image (panel~(a)) is
depicted along with a transformed version $\Trafo{\cdot}{\Pert}$ with
$\Pert \in\Pertset$ (panel~(b)) and the respective predictions by a standard neural network classifier.

\section{Experimental setup}\label{sec:experimental_setup}

In our experiments, we compare invariance-inducing regularization
incorporated via various augmentation-based methods (as described in
Section~\ref{sec:diffreg}) used on standard networks and
representative spatial equivariant networks trained using standard
optimization procedures.
\subsection{Spatial equivariant networks}
\label{sec:spatnet}
We compare the robust prediction accuracies from
networks trained with the regularizers with three specialized architectures, designed to be
equivariant against spatial transformations and translations:
(a) \gresnet\:\:(\grn) \cite{Cohen16} using p4m convolutional layers (90 degree rotations, translations and mirror reflections) on CIFAR-10;
(b) Equivariant Transformer Networks (ETN) \cite{Tai19}, a generalization of Polar Transformer Networks (PTN) \cite{Esteves18}, on SVHN; 
and (c) Spatial Transformer Networks (STN) \cite{Jaderberg15} on SVHN.
A more comprehensive discussion of the literature on equivariant
networks can be found in
Sec.~\ref{sec:related}. 
We choose the architectures listed above based on
availability of reproducible code and previously reported state-of-the art
standard accuracies on SVHN and CIFAR10.
We train \grn, \stn\,and \etn\, using standard augmentation as described in
Sec.~\ref{sec:exp_details} (\std) and random rotations in addition (\aug{\std}). 
Out of curiosity we also trained a ``two-stage'' STN where we train
the localization network separately in a supervised fashion. Specifically, we use a randomly transformed version of the training data, treating the transformation parameters as prediction targets. 
Details about the implementation and results can be found in Sec.~\ref{sec:twostage} in the \suppmat.

 

\subsection{Transformations}
The transformations that we
consider in Sec.~\ref{sec:results} are small rotations (of up to
$30^\circ$) and translations in two dimensions of up to 3 px
corresponding to approx.\ 10\% of the image size.  For augmentation
based methods we need to generate such small transformations for a
given test image.  Although the definition of a transformation
$\Trafo{X}{\Pert}$ in the theoretical section using the corresponding
continuous image functions is clean, we do not have acccess to the
continuous function in practice since the mapping is in general not
bijective.  Instead, we use bilinear interpolation, as implemented in
TensorFlow and in a differentiable version of a transformer \cite{Jaderberg15}
for first order attack and defense methods.

\begin{figure}[htbp]
\begin{minipage}[c]{0.45\textwidth}
  \begin{center}
    \begin{minipage}[t]{0.3\hsize}
      \vspace{0pt}
      \includegraphics[width=\textwidth]{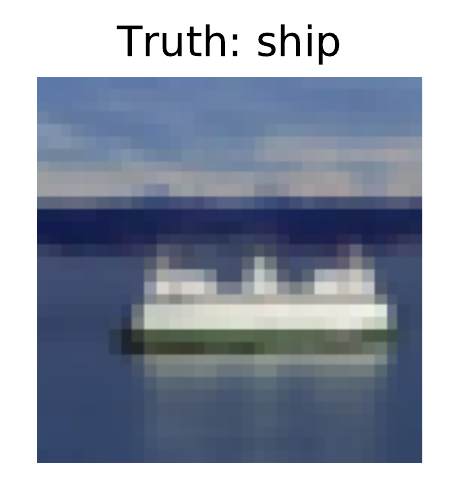}
    \end{minipage}
    \begin{minipage}[t]{0.3\hsize}
      \vspace{0pt}
      \includegraphics[width=\textwidth]{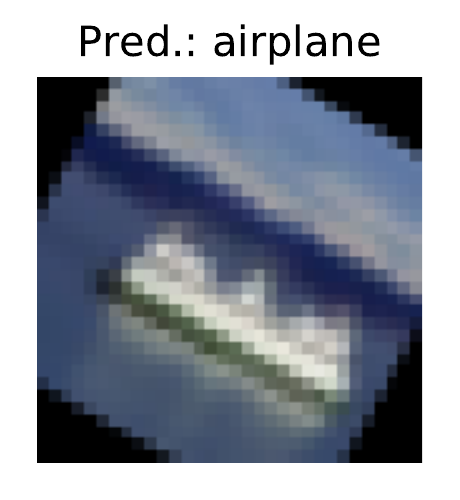}
    \end{minipage}

    \begin{minipage}[t]{0.3\hsize}
      \begin{center}
        \small (a)
      \end{center}
    \end{minipage}
    \begin{minipage}[t]{0.3\hsize}
      \begin{center}
        \small (b)
      \end{center}
    \end{minipage}
  \end{center}
\end{minipage}
\begin{minipage}[c]{0.55\textwidth}
\vskip -0.1in
\caption{\small Example images and classifications by the Standard model.
(a) An image that is correctly classified for most of the rotations in the considered grid.
  (b) One rotation for which the image shown in (b) is misclassified as ``airplane''. \label{fig:ex_trafo}
}\end{minipage}
\end{figure}

On top of interpolation, rotation also creates edge artifacts at
the boundaries, as the image is only sampled in a bounded set.
The empty space that
results from translating and rotating an image is filled with black
pixels (\emph{constant padding}) if not noted otherwise. Fig.~\ref{fig:ex_trafo} (b) shows an example.
\cite{Engstrom17} additionally
analyze a ``black canvas`` setting where the images are padded with
zeros prior to applying the transformation, ensuring that no
information is lost due to cropping. Their experiments show that the
reduced accuracy of the models cannot be attributed to this
effect. Since both versions yield similar results, we report results
on the first version of pad and crop choices, having input images of the same size as the
original.

\subsection{Attacks and defenses}\label{sec:attacks_defenses} 

The attacks and defenses we choose essentially follow the setup in
\cite{Engstrom17}. The \textit{defense} refers to the procedure at
training time which aims to make the resulting model robust to
adversarial examples. It generally differs from the (extensive)
\textit{attack} mechanism performed at evaluation time to assess the
model's robustness due to computational constraints.

\textbf{Considered attacks} First order methods such as projected
gradient descent that have proven to be most effective for
$\ell_\infty$ transformations are not optimal for finding adversarial
examples with respect to rotations and translations. In particular,
our experiments confirm the observations reported in~\cite{Engstrom17}
that the most adversarial examples can be found through a {\bf grid
  search}.
For the grid search attack, the compact
perturbation set $\Pertset$ is discretized to find the transformation
resulting in a misclassification with the largest loss $\loss$. In contrast to the case of
$\ell_\infty$-adversarial examples, this method is computationally
feasible for the 3-dimensional spatial parameters.
We consider a default grid of 5 values per translation direction and
31 values for rotation, yielding 775 transformed examples that are
evaluated for each $X_i$. We refer to the accuracy attained under this
attack as \textit{grid accuracy}. How did we ensure the number of
transformations in the grid are sufficient? Considering
with a finer grid of 7500 transformations for a subset of the experiments, summarized in Table~\ref{tab:finer_grid}, showed only  minor reductions
in accuracy compared to the coarser grid. Therefore, we chose the latter for computational reasons.

\textbf{Considered defenses} 
For the adversarial example which maximizes
either the loss or regularization function, we use the following
defense mechanisms:

\begin{itemize}
\item \textbf{$\wok$:} 
  At every iteration $t$, we sample $\numwok$ different perturbations for
  each image in the batch. The one resulting in the highest function
  value is used as the maximizer.  Most of our experiments are
  conducted with $k=10$ consistent with~\cite{Engstrom17} as a higher
  $k$ only improved performance minimally (see Table~\ref{tab:compareadv3}).

\item \textbf{Spatial PGD:} In analogy to
common practice for $\ell_p$ adversarial training as in
e.g.~\cite{Szegedy13, Madry18}, 
the \fo\, mechanism uses projected gradient descent with respect
to the translation and rotation parameters with projection
on the constrained set $\Pertset$ of transformations.
We consider 5 steps of PGD, starting from a random initialization,
with step sizes of $[0.03, 0.03, 0.3]$ (following \cite{Engstrom17})
for horizontal-, vertical translation and rotation respectively. A
discussion on the discrepancy between S-PGD as a defense and attack
mechanism can be found in Section~\ref{sec:moreresults}.

\item \textbf{Random:} Data augmentation with a distinct random perturbation per image and iteration. This can be seen as the most naive ``adversarial'' example as it corresponds to $\wok$ with $k=1$.
\end{itemize}


\subsection{Training details}\label{sec:exp_details}



The experiments are conducted with deep
neural networks as the function space $\Fspace$ and $\ell$ is the cross-entropy loss.  
In the main paper we consider the datasets SVHN \cite{Netzer11} and
CIFAR-10 \cite{Krizhevsky09}. For the non-specialized architectures,
we train a ResNet-32 \cite{He16}, implemented in TensorFlow
\cite{Abadi2015}. For the Transformer networks STN and ETN
we use a 3-layer CNN as localization according to the default settings in the provided code of
both networks for SVHN and rot-MNIST.
For a subset of the experiments we also report
results for CIFAR-100 \cite{Krizhevsky09}
in the \suppmat.

We train the baseline models with standard data augmentation:
random left-right flips and random translations of $\pm 4\px$ followed
by normalization.  Below we refer to the models trained in this
fashion as ``\std''.  For the models trained with one of the defenses
described in Sec.~\ref{sec:attacks_defenses}, we only apply random
left-right flipping since translations are part of the adversarial
search. The special case of data augmentation (with translations and
rotations, i.e.\ the defense ``random'') without regularization is
refered to as $\aug{\std}$. 

For optimization of the empirical training loss, we run standard
minibatch SGD with a momentum term with parameter $0.9$ and weight
decay parameter $0.0002$.  We use an initial learning rate of $0.1$
which is divided by $10$ after half and three-quarters of the training
steps. Independent of the defense method, we fix the number of
iterations to $80000$ for SVHN and CIFAR-10, and to $120000$ for
CIFAR-100. For comparability across all methods, the number of unique
original images in each iteration is $64$ in all cases. For the
baselines $\std, \aug{\std}$ and $\at$, we additionally trained with a
conventional batch size of $128$ and report the higher accuracy of
both versions. For the regularized methods, the value of
  $\lambda$ is chosen based on the test grid accuracy. All models are
trained using a single GPU on a node equipped with an NVIDIA GeForce
GTX 1080 Ti and two 10-core Xeon E5-2630v4 processors.

\begin{figure}[htbp]
   \begin{center}
       \includegraphics[width=.45\textwidth]{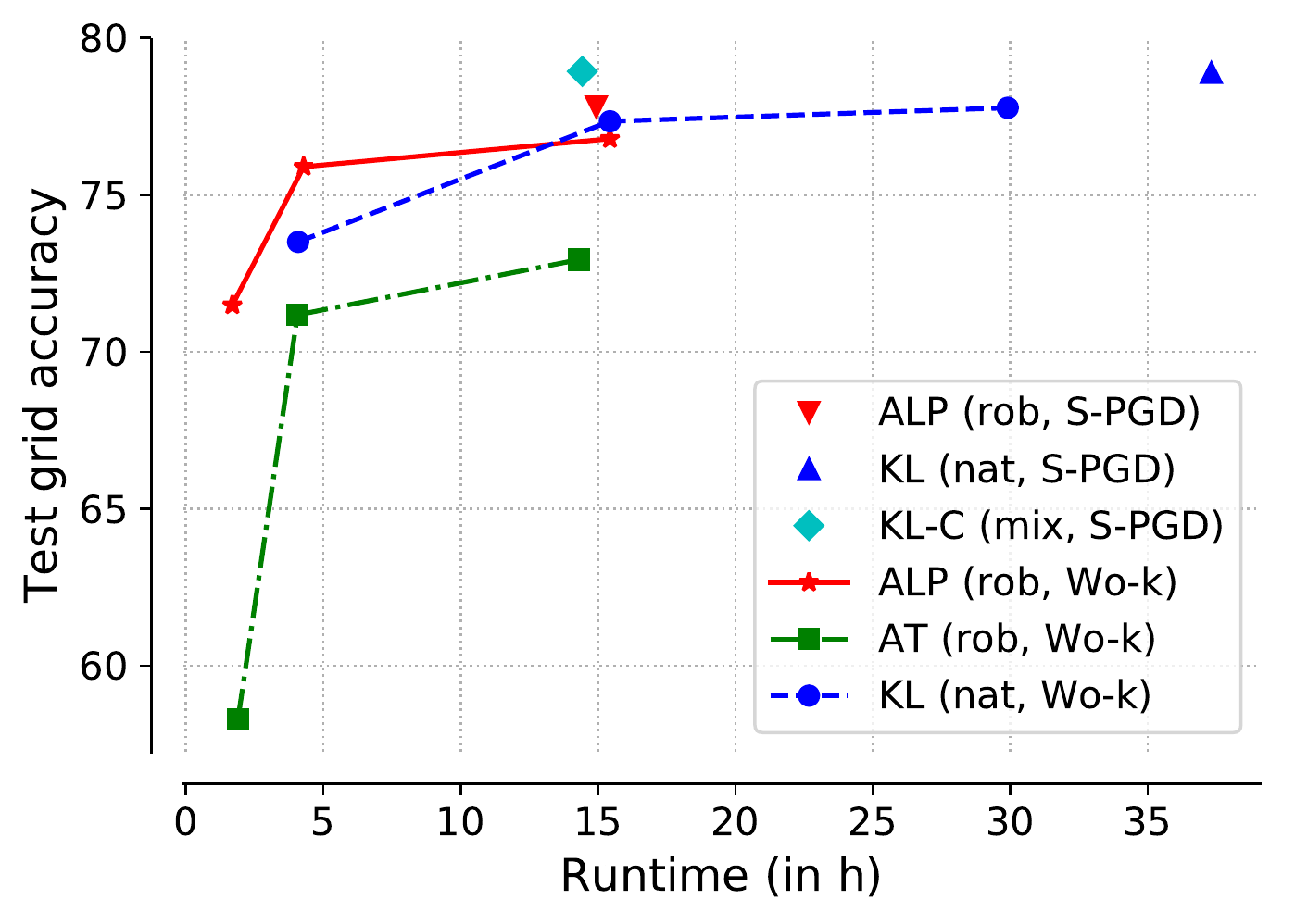}
   \end{center}
     \caption{\small Mean runtime for different methods on CIFAR-10. The connected points correspond to \wokabb\, defenses with $\numwok \in \{1, 10, 20\}$.  \label{fig:runtime} 
     }
\end{figure}

\section{Empirical Results}
\label{sec:results}

We now compare the {natural test accuracy} (standard accuracy on the
test set, abbreviated as \emph{nat}) and test grid accuracy (as
defined in Sec.~\ref{sec:attacks_defenses}, abbreviated as \emph{rob})
achieved by standard and regularized (adversarial) training techniques
as well as specialized spatial equivariant architectures described in
Sec.~\ref{sec:spatnet}.  For clarity of presentation, the naming
convention we use in the rest of the paper consists of the following
components: (a) \regtype: refers to what regularizer was used (\AT,
\ALP, \ltwo, \tr, or \trcce \: as defined in
Section~\ref{sec:diffreg}); (b) \batchtype: indicates whether the
gradient of the loss is taken with respect to the adversarial examples
(\rob), natural examples (\nat) or both (\mix), and (c) \defvar: the
mechanism used to find the adversarial example, including random
(\rnd), \wo{k} (\woabb{k}) and spatial PGD (\fo) as described in
Sec.~\ref{sec:attacks_defenses}.  Thus, \regtype (\batchtype, \defvar)
corresponds to using \regtype as the regularization function, the
examples defined by \batchtype\; in the gradient of the loss and the
defense mechanism \defvar\, to find the augmented or adversarial
examples.

In Table~\ref{tab:regspat}, we report results for a subset of the \regtype (\batchtype, \defvar) combinations to facilitate comparisons. Tables with many more combinations can be found Tables~\ref{tab:compareadv1}--\ref{tab:compareadv5} in the \suppmat.
We report averages (standard errors are contained in Tables~\ref{tab:compareadv1}--\ref{tab:compareadv5}) computed over five training runs with identical
hyperparameter settings.  
We compare all
methods by computing absolute and relative error reductions (defined
as $\frac{\text{absolute error drop}}{\text{prior error}}$). It is
insightful to present both numbers since the absolute values vary
drastically between datasets. 

\begin{table*}[t]
  \caption{Mean accuracies of models trained with various forms of regularized adversarial training as well as standard augmentation techniques (top) and spatial equivariant networks (bottom). $\star$ denotes standard augmentation plus random rotations. The highest accuracies per row are bolded.\label{tab:regspat}}
\vskip 0.15in
       {\setlength{\tabcolsep}{4pt}
\begin{tabular}{c c c c : c c c : c c}
  & \std & \stdaug
  & \makecell{\AT\, (\rob, \\\woabb{10})} & \makecell{\tr \; (\rob, \\ \woabb{10})} & \makecell{\ltwo\;(\rob, \\ \woabb{10})}
  & \makecell{\ALP \;(\rob, \\ \woabb{10})} &  \makecell{\trcce \;(\mix, \\ \fo)} & \makecell{\ALP \;(\rob, \\ \fo)} \\
  \hline
  SVHN (nat) & 95.48 & 93.97 & 96.03 & 96.13  & {\bf 96.53} & 96.30 & 96.14 & 96.11  \\
  \qquad\: (rob) & 18.85 & 82.60 & 90.35 & {\bf 92.71} & 92.55 & 92.04 & { 92.42}  & 92.32\\
  \hdashline
  CIFAR (nat) & {\bf 92.11} & 89.93 & 91.76 & 90.41 & 90.53 & 90.11 & 89.98 & 89.85 \\
  \qquad\quad (rob) & 9.52 & 58.29 & 71.17 & { 77.47} & 77.06 & 75.9 & {\bf 78.93} & 77.80\\
  \hline
  \hline
   \end{tabular}}
   {\setlength{\tabcolsep}{5.5pt}
  \begin{center}
  \begin{tabular}{c c c c c c c }
               & \small{\grn} & \small{\aug{\grn}} & \etn & \aug{\etn} & \stn & \aug{\stn} \\
  \hline
  SVHN (nat) &   96.07 & 95.05 & 95.53 & 95.57 & 95.61 & 95.55 \\
  \qquad\: (rob)  & 25.12 & 84.9 & 13.15 & {\bf 84.21} & 36.68 & 79.28 \\
  \hdashline
  CIFAR (nat)  & 93.39 & 93.08 & -- & -- & -- & --  \\
  \qquad\quad (rob)  & 16.85 & {\bf 71.64} & -- & -- & -- & -- \\
  \hline
  \end{tabular}
  \vspace{-0.05in}
  \end{center}}
\end{table*}

\begin{table*}[t]
    \caption{Mean accuracies of models trained with various forms of regularized adversarial training. Left: All adversarial examples were found via $\woabb{10}$; right: unregularized ($\stdaug$) and regularized data augmentation where the optimum is bolded for each row.\label{tab:comparereg}}
  \begin{minipage}[c]{\textwidth}
 \begin{center}
\vskip 0.15in
   {\setlength{\tabcolsep}{4pt}
     \begin{tabular}{c c c c | c c  c c c}  
               & \makecell{\tr \; (\nat, \\ \woabb{10})} &  \makecell{\ltwo \; (\nat, \\ \woabb{10})} & \makecell{\ALP \; (\nat, \\ \woabb{10})} & \stdaug & \makecell{\ltwo \; (\nat, \\ \rnd)} & \makecell{\tr \; (\nat, \\ \rnd)}  & \makecell{\ltwo \; (\rob, \\ \rnd)}  &  \makecell{\tr \; (\rob, \\ \rnd)}  \\
  \hline
  SVHN (nat) &   96.00 & 96.05 & 96.39   & 93.97 &{\bf 96.34} &  96.16 & 96.09 & 96.23\\
  \qquad\: (rob)  & 92.27 & 92.16 &  91.98& 82.60 &  90.51 & 90.69  & 90.48 &  {\bf 90.92} \\
  \hdashline
  CIFAR (nat)  & 90.83 & 88.32 &  88.78 & {\bf 89.93}& 87.80 & 89.33  & 88.75 &  89.47  \\
  \qquad\quad (rob)  & 77.34 &  75.64 &  75.43   & 58.29&  71.60  & {\bf 73.50} & 71.49 &  73.22 \\
  \hline
   \end{tabular}}
 \end{center}
  \end{minipage}
\end{table*}

\textbf{Effectiveness of augmentation-based invariance-inducing
  regularization} In Table~\ref{tab:regspat} (top), the three leftmost
columns represent unregularized methods which all perform worse in
grid accuracy than regularized methods and the two right-most columns
represent adversarial examples with respect to the classification
cross entropy loss found via S-PGD. When considering the three
regularizers (\tr, \ltwo, \ALP) with the same \batchtype\, and
\defvar\, (here chosen to be ``\rob'' and \woabb{10}) regularized
adversarial training improves the grid accuracy from $71.17\%$ to
$77.47\%$ on CIFAR-10 and $90.35\%$ to $92.71\%$ on SVHN,
corresponding to a relative error reduction of $22\%$ and $24\%$
respectively. The same can be observed when comparing data
augmentation $\stdaug$ and its
regularized variants $\ltwo(\cdot,\rnd), \tr(\cdot,\rnd)$ in
Table~\ref{tab:comparereg}. Together with Tables~\ref{tab:compareadv2}
and \ref{tab:compareadv3}, S-PGD seems to be the more efficient
defense mechanism compared to $\wok$ even when $k$ is raised to $20$,
with comparable computation time.

\textbf{Computational considerations}
In Figure~\ref{fig:runtime}, we plot the grid accuracy vs.\ the
runtime (in hours) for a subset of regularizers and defense mechanisms
on CIFAR-10 for clarity of presentation.  How much overhead is needed
to obtain the reported gains?  Comparing \AT(\rob, \wokabb) (green line) and
\ALP(\rob, \wokabb) (red line) shows that significant improvements in grid
accuracy can be achieved by regularization with only a small
computational overhead.
What if we make the defense stronger? 
While the leap in robust accuracy from $\woabb{1}$ (also referred to as
\rnd) to \woabb{10} is quite large, increasing $k$ to 20 only gives
diminishing returns while requiring $\sim 3\times$ more training time.
This observation is summarized exemplarily for both \tr\ and \alp\ 
regularizer on CIFAR-10 in Table~\ref{tab:wok_diffk}.
Furthermore, for any fixed training
time, regularized methods exhibit higher robust accuracies where the
gap varies with the particular choice of regularizer and defense mechanism.

\textbf{Comparison with spatial equivariant networks} Although the
rotation-augmented \gresnet\, obtains higher grid (SVHN: $84.9\%$,
CIFAR-10: $71.64\%$) and natural accuracies (SVHN: $95\%$, CIFAR-10:
$93.08\%$) than the rotation-augmented Resnet-32 on both SVHN (grid:
$82.60\%$, nat: $93.97\%$) and CIFAR10 (grid: $58.29\%$, nat:
$89.93\%$), regularizing standard data augmentation
(i.e.\ regularizers with ``rnd'', see Table~\ref{tab:comparereg}
(right)) using both the $\ltwo$ distance and the $\tr$ divergence
matches the \gresnet\, on CIFAR10 (\ltwo: $71.60\%$, \tr: $73.50\%$)
and surpasses it on SVHN on grid (\ltwo: $90.51\%$, \tr: $90.69\%$)
and natural accuracies by a relative grid error reduction of
$\sim37\%$. The same phenomenon is observed for the augmented
\etn\,and \stn\,on SVHN.\footnote{We had difficulties to train both
  \etn\,and \stn\,to higher than $86\%$ natural accuracy for CIFAR10
  even after an extensive learning rate and schedule search so we do
  not report the numbers here.}  In conclusion, regularized
augmentation based methods match or outperform representative
end-to-end networks handcrafted to be equivariant to spatial
transformations. 




\textbf{Trade-off natural vs. adversarial accuracy} SVHN is one of the
main datasets (without artificial augmentation like in
rot-MNIST~\cite{larochelle07}) where spatial equivariant networks have
reported improvements on natural accuracy.  This is due to the
inherent orientation variance in the data.  In our mathematical
framework, this corresponds to the assumption in
Theorem~\ref{theo:tradeoff} of the distribution on the transformation
sets having support $\SubGorbit{z}$.  Furthermore, as all numbers in
SVHN have the same label irrespective of small rotations of at most 30
degrees, the first assumption in Theorem~\ref{theo:tradeoff} is also
fulfilled. Table~\ref{tab:regspat} and ~\ref{tab:comparereg} confirm
the statement in the Theorem that improving robust accuracy may not
hurt natural accuracy or even improve it: For SVHN, adding
regularization to samples obtained both via $\woabb{10}$ adversarial search or random transformation (\rnd) consistently not only helps robust but also standard
accuracy.

\textbf{Comparing the effects of different regularization parameters
  on test grid accuracy}
We study Tables~\ref{tab:regspat} and~\ref{tab:comparereg} and attempt
to disentangle the effects by varying only one parameter.  For example
we can observe that, computational cost aside, fixing any regularizer
defense to $\woabb{10}$, the robust regularized loss \regtype(\rob,
$\woabb{10}$) (i.e., $\PopLoss{\rob}(f;\Reg)$) does better (or not
statistically significantly worse) than \regtype(\nat, $\woabb{10}$)
(i.e., $\PopLoss{\nat}(f;R)$). Furthermore, the KL regularizer
generally performs better than $\ltwo$ for a large number of
settings. A possible explanation for the latter could be that $\KL$
upper bounds the squared $\ell_2$ loss on the probability simplex and
is hence more restrictive.

\textbf{Choice of $\lambda$}
The different regularization methods peak at
different $\lambda$ in terms of grid accuracy. However, they outperform unregularized methods in a large range of $\lambda$ values, suggesting that
well-performing values of $\lambda$ are not difficult to find in
practice. These can be seen in Figures~\ref{fig:grid_nat_overlambda_rnd} and~\ref{fig:grid_nat_overlambda_wo} in the \suppmat.
 

There are many more interesting experiments we have conducted for
subsets of the defenses and datasets illustrating different
phenomena that we observe. For example we have analyzed a finer grid for the grid search attack and evaluated $\fo$ as an attack mechanism. A detailed
discussion of these experiments can be found in
Sec.~\ref{sec:moreresults}.

\vspace{-0.2in}
\section{Related work}
\label{sec:related}
\textbf{Group equivariant networks} There
are in general two types of approaches to incorporate spatial
invariance into the network. In one of the earlier works in the neural
net era, Spatial Transformer Networks were introduced
\cite{Jaderberg15} which includes a transformer module that
predicts transformation parameters followed by a transformer. Later
on, one line of work proposed multiple filters that are discrete group
transformations of each other \cite{Pollefeys16, Marcos17, Cohen16,
  Zhou17, Worrall17}.  For continuous transformations,
steerability \cite{Weiler18, Cohen18}
and coordinate transformation \cite{Esteves18, Tai19} based approaches
have been suggested. Although these approaches have resulted
in improved standard accuracy performances, it has not been
rigorously studied whether or by how much they improve upon
regular networks with respect to robust test accuracy.

\noindent \textbf{Regularized training} Using penalty regularization to
encourage robustness and invariance when training neural networks
has been studied in different contexts: for
distributional robustness \cite{Heinze17}, domain generalization
\cite{Motiian17}, $\ell_p$ adversarial training \cite{Na18, Kannan18,
  Zhang19}, robustness against simple transformations \cite{Cheng19}
and semi-supervised learning \cite{Zheng16, Xie19}.
These approaches are based on augmenting the training data either
\emph{statically} \cite{Heinze17, Motiian17, Cheng19, Xie19}, ie. before fitting the
model, or \emph{adaptively} in the sense of adversarial training, with different
augmented examples per training image generated in every
iteration \cite{Kannan18, Na18, Zhang19}.

\noindent \textbf{Robustness against  simple transformations}
Approaches targeting adversarial accuracy for
simple transformations have used attacks and defenses in the spirit
of PGD (either on transformation space \cite{Engstrom17} or on input
space projecting to transformation manifold \cite{Kanbak18}) and
simple random or grid search \cite{Engstrom17, Pei17}.
Recent work~\cite{Dumont18} has also evaluated some
rotation-equivariant networks with different training and attack
settings which reduces direct comparability with e.g. adversarial
based defenses~\cite{Engstrom17}.

\section{Conclusion}
In this work, we have explored how regularized augmentation-based
methods compare against specialized spatial equivariant networks in
terms of robustness against small translations and
rotations. Strikingly, even though augmentation can be applied to
encourage \emph{any} desired invariance, the regularized methods adapt
well and perform similarly or better than specialized networks.
Furthermore, we have introduced a theoretical framework incorporating
many forms of regularization techniques that have been proposed in the
literature. Both theoretically and empirically, we showed that for
transformation invariances and under certain practical assumptions on
the distribution, there is no trade-off between natural and
adversarial accuracy which stands in contrast to the debate around
$\ell_p$-perturbation sets. In summary, it is advantageous to replace
unregularized with regularized training for both augmentation
and adversarial defense methods. With regard to the regularization
parameter choice we have seen that improvements can be
obtained for a large range of $\lambda$ values, indicating that this
additional hyperparameter is not difficult to tune in practice.  In
future work, we aim to explore whether specialized
architectures can be combined with regularized adversarial training to
improve upon the best results reported in this work.

\section{Acknowledgements}
We thank Ludwig Schmidt for helpful discussion, Nicolai Meinshausen
for valuable feedback on the manuscript and Luzius Brogli for initial
experiments.  FY was supported by the Institute for Theoretical
Studies ETH Zurich, the Dr. Max R\"ossler and Walter Haefner
Foundation and the Office of Naval Research Young Investigator Award
N00014-19-1-2288.

\bibliography{references}
\bibliographystyle{plain}

\appendix
\clearpage
\onecolumn
\section{Appendix}

\subsection{Rigorous definition of transformation sets and choice of $\Pertset$}
\label{sec:groups}


In the following we introduce the concepts that are needed to
rigorously define transformation sets that are subsets of the
finite-dimensional (sampled) image space $\X \subset \RN^d$.  In particular, because
rotations of continuous angles are not well-defined for sampled images
we need to introduce the space of image functions $\Gspace$ with
elements $I:\RN^2 \to [0,255]^3$, i.e.\ $I$ maps Euclidean coordinates
in $\RN^2$ to the RGB intensities of an image.  The observed
finite-dimensional vector is then a sampled version of an image
function $I$. Here we assume that the sampling operator is bijective,
with rigorous definitions later in the section.

Next we define subsets in the continuous function space and then
transfer the concept back to the finite-dimensional $\X$.
Let us define the
symmetric group $\Group$ of all rotations and horizontal and vertical
translations acting on $\Gspace$. 
We denote the elements in the group by $g_\Pert$, uniquely parameterized
by $\Pert \in \RN^3$ and can be represented by a coordinate transform
matrix $G_\Pert$, see e.g.~\cite{Cohen16}. Two of the three dimensions
represent the values for the translations and the third represents the
rotation.



The transformed image (function) $g_\Pert(I) \in \Gspace$ can be
expressed by $g_\Pert(I)(v) = I(G_\Pert^{-1}v)$ for each $v\in \RN^2$
where $G_\Pert$ is the coordinate transform matrix associated with
$\Pert \in \RN^3$ as in \cite{Cohen16}. For each $I\in \Gspace$, the
group orbit is $\Gorbit{I} \defn \{g_\Pert(I) :
g_\Pert\in \Group\}$. By definition, the group orbits partition the
space $\Gspace$ and every $I \in \Gspace$ belongs to a unique orbit.
 
\vspace{-0.2in}
\paragraph{Subsets of orbits}
In our setting, requiring invariance in the entire orbit
(i.e.\ with respect to all translations and rotations) is too
restrictive. First of all, large transformations rarely occur in nature
because of physical laws and common human perspectives (an upside down
tower for example). Secondly, in image classification, robustness is
usually only required against adversarial attacks which would not fool
humans, i.e.\ lead them to mislabel the image. If the transformation
set is too large, this requirement is no longer fulfilled.
For this purpose we consider a closed subset
$\SubGroup{I}$ of each group orbit.
It follows from the group orbit definition that for every $I$ it
either belongs to one unique or no such set.

\begin{figure}[!htp]
\begin{center}
\begin{tabular}{ccc}
\hspace{-0.8cm}
\includegraphics[scale=0.42, keepaspectratio=true, trim={0 0 0 0}]{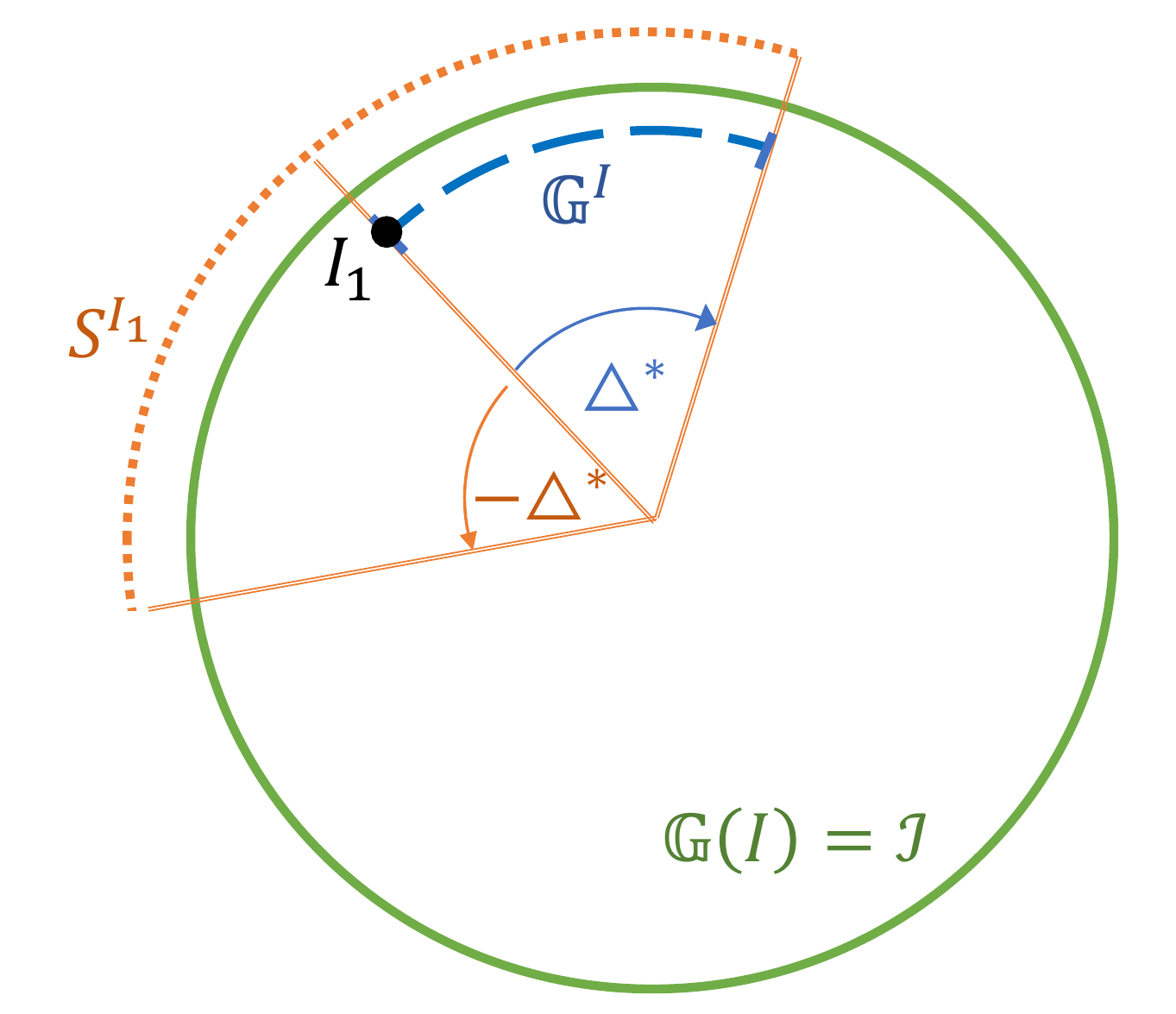} &
\hspace{-0.8cm}
\includegraphics[scale=0.42, keepaspectratio=true, trim={0 0 0 0}, clip]{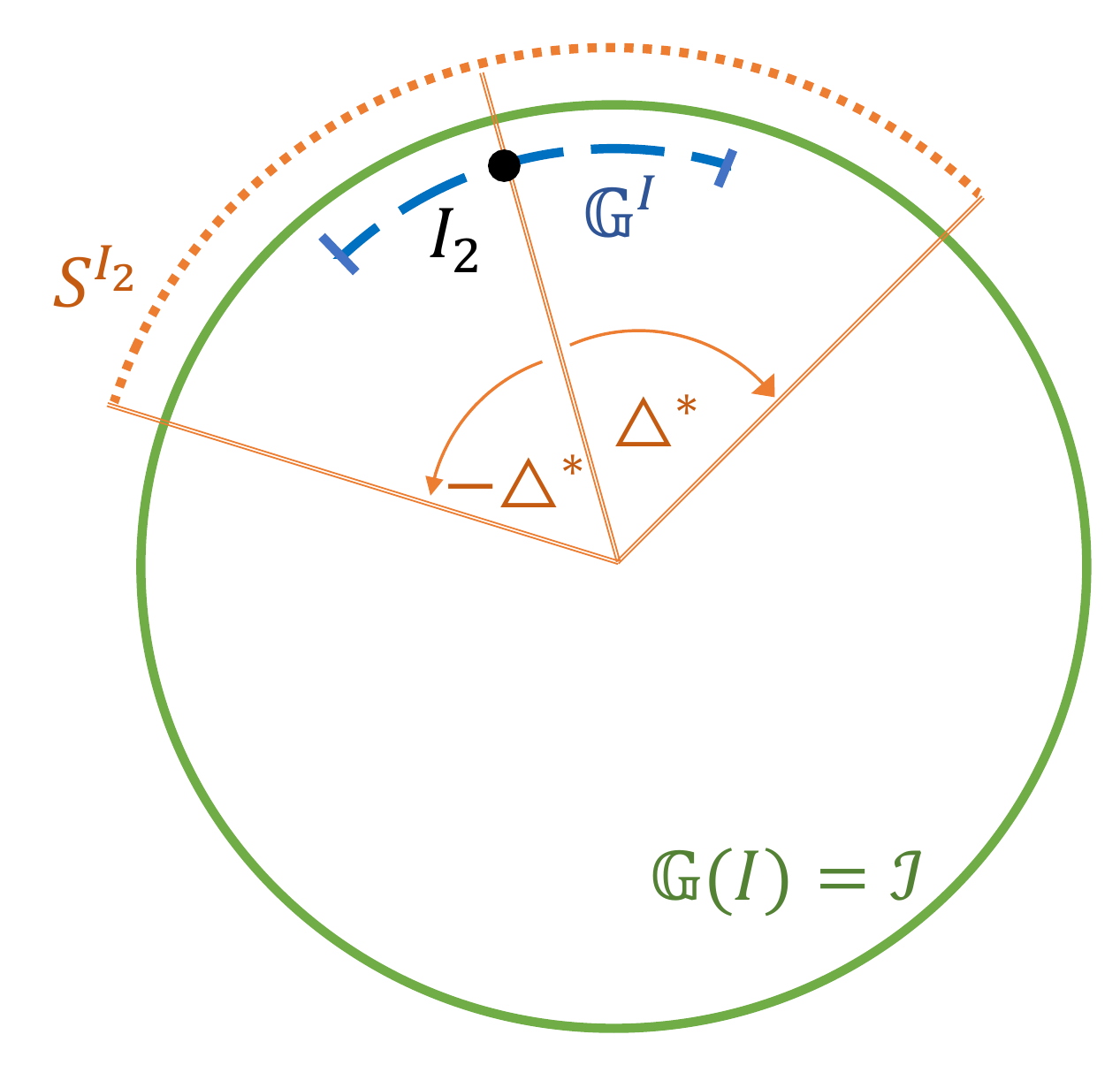} \hspace{0cm}&
\includegraphics[scale=0.42, keepaspectratio=true, trim={0 0 0 0}, clip]{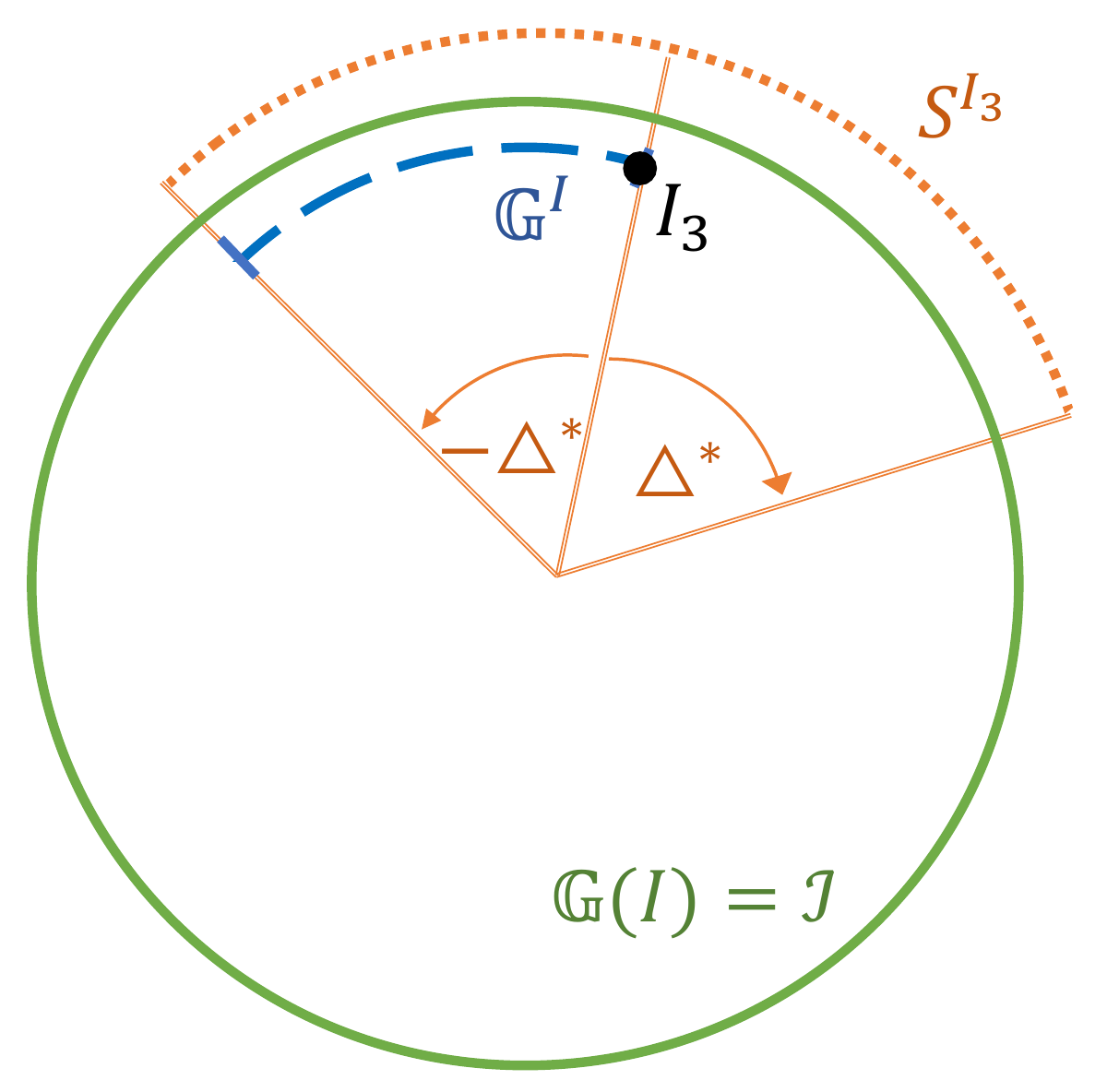}
\\
(a) & (b) & (c)
\end{tabular}
\caption{Illustration of an example where one group orbit $\Group(I) =: \Gspace$ is the entire space of images and $I$ is an arbitrary image in the orbit $\Group(I)$. We depict one subset of the orbit $\SubGroup{I}$ and the effective search sets $\searchset{I'}$ for different instantiations $I'\in\SubGroup{I}$ defined by the transformation search set $\Pertset$: (a) $I_1$ on the left boundary of $\SubGroup{I}$, (b) $I_2$ in the interior of $\SubGroup{I}$ and (c) $I_3$ on the right boundary of $\SubGroup{I}$. The effective search sets are centered around each instantation $I_j$. The necessity of symmetry of the minimal set of transformations $\Pertset$ arises from the requirement to cover $\SubGroup{I}$ from both boundary points and the maximum transformation vector $\Pertdiam$ that defines $\Pertset = (-\Pertdiam, \Pertdiam)$ is determined by the maximum transformation in $\SubGroup{I}$ (in blue). 
\label{fig:orbits}}
\end{center}
\end{figure}

As described in the paragraph of Equation~\eqref{eq:advminapp}, when
observing a (sampled) image $I'$ in the training set, we do not know
where in its corresponding subset $\SubGroup{I'}$ it lies. At the same
time, for our augmentation-based methods, we do not want the set
$\Pertset$ of transformations that we search over
(\emph{transformation search set} for short), to be image
dependent. Instead, in this construction we aim to find $\Pertset$ to
be the smallest set of transformations such that~\eqref{eq:advminapp}
is satisfied. For this purpose, it suffices that the effective search
set of images $\searchset{I'}$ for any image $I' \in \SubGroup{I}$
covers the corresponding subset $\SubGroup{I}$ for all $I$, i.e.
\begin{equation*}
\searchset{I'} \defn \{g_{\Pert}(I'): \Pert \in \Pertset\} \supset \SubGroup{I}.
\end{equation*}
Here we give an explicit construction of $\Pertset$ using
the maximal transformation for each subset $\SubGroup{I}$ that is
needed to transform an image of the subset to another.
In particular, we define the maximal transformation vector $\Pertdiam \in \RN^3$ by the element-wise maximum over all such maximum transformations
\begin{equation*}
(\Pertdiam)_j \defn \max_{I\in\Gspace} \max_{U,U' \in\SubGroup{I}} |(\Pert)_j| \text{ s.t. } U' = g_{\Pert}(U)
\end{equation*}
for $j=1, \dots, 3$. Although the subsets themselves for each image are
not known, using prior knowledge in each application one can usually
estimate the largest possible range of transformations $\Pertdiam$
against which robustness is desired or required. For example for images,
one could use experiments with humans to determine for which range of angles
their reaction time to correctly label each image stays approximately constant.
The maximal vector $\Pertdiam$ can now be used to
determine the minimal set of transformations $\Pertset =
(-\Pertdiam, \Pertdiam)$.
A simplified illustration for when $\Gspace$ consists of just one
orbit (corresponding for example to one image function and all its
rotated variants) can be found in Figure~\ref{fig:orbits}.

\paragraph{Sampling issues}
In reality, the observed image is not a function on $\RN^2$ but a
vector $z \in \RN^{w\times h}$ that is the result of sampling an image
function $I \in \Gspace$. We use $\Phi$ to denote the
sampling operator and hence $z = \Phi(I)$. Then the space of
observed finite dimensional images $\X$ is the range space of $\Phi$.
In order to counter the problem that the sampling operator is in
general not injective, we add another constraint to $\Gspace$ by
requiring that $\Phi$ is bijective so that the quantity $I_z
= \Phi^{-1}(z)$ is well-defined. That is, for a finite-dimensional
image $z\in \X$, there exists exactly one possible continuous image
$I_z \in \Gspace$.  As a consequence, if $z$ and a transformed version
$z'$ exist in $\X$, then $I_z = I_{z'}$.  This is a rather technical
assumption that is typically fulfilled in practice.
In the main text, we also refer to $\Trafo{z}{\Pert}
= \Phi(g_\Pert(I_z)) \in \X$ as the image corresponding to the sampled
image $z$ transformed by the group element $g_\Pert$.



We can now define specific $\SubGroup{I}$ to be the subsets of $\Group(I)$ such
that with $z = \Phi^{-1}(I)$, the set $\SubGorbit{z} = \{\Phi(I): I \in \SubGroup{I_z}\}$ corresponds to the support of the marginal distribution $\P$ on $\{\Phi(I): I\in\Group(I_z)\}$. We refer to $\SubGorbit{z}$ as transformation sets.
By definition of $\SubGroup{I}$ and bijectivity of $\Phi$, there is an
injective mapping from any $z \in \X$ to the set of transformation sets $\SetOrbit$.

\subsection{Proof of Theorem~\ref{theo:1}}
\label{sec:theo1proof}
Please refer to Section~\ref{sec:groups} for the necessary notation for this section.
Furthermore, define  $\PopLoss{\nat}(f)\defn \PopLoss{\nat}(f; 0,0)$.

We prove the first statement of the theorem by contradiction. Let
$\frob$ be the minimizer of $\PopLoss{\rob}(f)$ and let us assume that
$\frob \not \in \InvF$ and in particular that it is constant on all
transformation sets except $\SubGorbit{z} \in \SetOrbit$ and the
marginal distribution over $\SetOrbit$ that can be defined as
$P(\{\SubGorbit{X} = U\}) = P(\{X\in U\})$ for any $U\in\SetOrbit$,
is discrete (for simplicity of presentation) and $\SubGorbit{z}$ has
non-zero probability.

Let's assume that there is at least one transformation set $\SubGorbit{z}$,
on which $\frob$ is not constant and collect all different values in
the set $A = \{ \frob(x): x\in \SubGorbit{z} \}$ (with cardinality
strictly bigger than $1$ since $f$ not constant) and denote the
distribution over $x\in\SubGorbit{z}$ by $P_z$.
Since there
is a unique mapping $\orbitfunc$ that maps each $x\in \X$ to a unique
transformation (see Section~\ref{sec:groups}), we can lower bound of
the robust loss as follows for any $z\in\X$:

\begin{align}
& \EE_{X,Y} \sup_{x' \in\SubGorbit{X}} \loss(f(x'),  Y)\nonumber \\
  = &\:\EE [\sup_{x' \in \SubGorbit{X}} \loss(f(x'), Y)|X \not\in \SubGorbit{z}] \P(\{X\not\in\SubGorbit{z} \}) + \EE_{Y|z}[ \sup_{x' \in \SubGorbit{z}} \loss(f(x'), Y)|\SubGorbit{z}] \P(\{X \in \SubGorbit{z}\}) \nonumber \\
  \geq &\:\EE [\sup_{x' \in \SubGorbit{X}} \loss(f(x'), Y)|X \not \in \SubGorbit{z}] \P(\{X\not \in \SubGorbit{z}\}) + \sup_{a\in A} \int \EE_{Y|x} \loss(a,Y) d P_z(x) \label{eq:prooflem}
\end{align}
where the inequality follows from
\begin{equation*}
\EE_{X|z} \EE_{Y|x} [\sup_{a\in A} \loss(a,Y) | X = x] |\SubGorbit{z} = \int \EE_{Y|x} \sup_{a\in A}\loss(a,Y) dP_z(x) \geq \sup_{a\in A} \int \EE_{Y|x} \loss(a,Y) dP_z(x).
\end{equation*}
The right hand side is minimized with respect to the set $A$ by
choosing $A = \{\abest\}$ where $\abest$ is defined as {$\abest = \argmin_a \int  \EE_{Y|x} \loss(a,Y) d
P_z(x)$} because setting $\fstar(x) = \abest$ for all $x\in \SubGorbit{z}$ and
$\fstar(x) = \frob(x)$ else leads to equality in equation~\eqref{eq:prooflem} and $\fstar \in \Fspace$ by assumption that $\InvF \subseteq \Fspace$. 
Morever, since $\P(\{X\in\SubGorbit{z}\}) > 0$ by assumption, choosing $\fstar(x)
= \abest$ for all $x\in\SubGorbit{z}$ implies
$\PopLoss{\rob}(\fstar)< \PopLoss{\rob}(\frob)$ which contradicts
optimality of $\frob$ and thus proves the first statement of the
theorem.

For the second statement let us  rewrite
\begin{align*}
\PopLoss{\rob}(f) &= \PopLoss{\nat}(f) + [\PopLoss{\rob}(f) - \PopLoss{\nat}(f)] \\
&=  \EE \loss(\func(X,\YID) +  \underbrace{\EE [\max_{\Pert' \in \Pertset} \loss(\func(\Trafo{X}{\Pert'}),\YID) - \loss(\func(X),\YID)]}_{\regadvnat(f)}
\end{align*}
By the first statement we know that the set of invariant functions that minimize the robust loss
\begin{equation*}
\Frob := \{f \in \InvF: \PopLoss{\rob}(f) \leq \PopLoss{\rob}(f') \quad \forall f' \in \Fspace\}
\end{equation*}
is non-empty. 
For all $f\in \Frob$, it holds by definition of $\InvF$ that 
$\regadvnat(f) = 0$.

Since $\InvF(\Reg) \subseteq \InvF$, the minimizers $\fmin$
of~\eqref{eq:regnat} satisfy
$\PopLoss{\nat}(\fmin) \leq \PopLoss{\nat}(f)$ for all
$f\in \InvF$. But because $\fmin$ in $\InvF$ we have $\PopLoss{\rob}(f)
= \PopLoss{\nat}(f)$ and it directly follows that $\fmin \in \Frob$.  The
same argument goes through for~\eqref{eq:regadv} since for all
$f \in \InvF$, we have $\PopLoss{\rob}(f) = \PopLoss{\nat}(f)$. This
concludes the proof of the theorem.

\subsection{Proof of Theorem~\ref{theo:tradeoff}}
\label{sec:proofcor}

On a high level, similar to the proof of Theorem~\ref{theo:1}, we can
construct a minimizer of the natural loss $\fstar \InvF$ given the
assumption that $Y \ortho X|\SubGorbit{z}$.
Since on $\InvF$ both losses are equivalent, together with
Theorem~\ref{theo:1} this shows that the robust minimizer also minimizes the
unconstrained natural loss.

Assume $\fnat \not \in \InvF$ minimizes $\PopLoss{\nat}(f)$, and in
particular, it is constant on all transformation sets except
$\SubGorbit{z}$ for some $z\in\X$.  Again by existence of a mapping
$\orbitfunc$ and by assumption  $Y \ortho X|\SubGorbit{z}$
we can write for any $f$
\begin{align}
\PopLoss{\nat}(f) &= \EE_{X, Y} \loss(f(X), Y) \label{eq:lnat}\\
&= \EE[\loss(f(X),Y) | X\not\in\SubGorbit{z}]\P(\{X\not \in \SubGorbit{z}\}) + \EE[\loss(f(X),Y) | \SubGorbit{z} ] \P(\{X \in \SubGorbit{z}\}) \nonumber\\
&= \EE[\loss(f(X),Y) | X\not\in\SubGorbit{z}]\P(\{X\not \in \SubGorbit{z}\}) +   \EE \big[ \EE_Y [\loss(f(X),Y)] | \SubGorbit{z} \big] \P(\{X \in \SubGorbit{z}\}).\nonumber
\end{align}
We then obtain 
\begin{align}
\EE \big[ \EE_Y [\loss(f(X),Y)] | \SubGorbit{z}\big] &= \int \EE[\loss(f(x),Y)|x] d P_z(x) \nonumber\\
&\geq \int \min_{x'\in\SubGorbit{z}} \EE[\loss(f(x'),Y)|x']  d P_z(x) = \EE \big[ \EE_Y [\loss (\fstar(X), Y)] | \SubGorbit{z}\big] \label{eq:ineq}
\end{align}
when setting
$\fstar(x)= \min_{x\in\SubGorbit{z}} \EE_Y\loss(\fnat(x),Y)
| \SubGorbit{z}$ for all $x\in\SubGorbit{z}$ and $\fstar(x)
= \fnat(x)$ otherwise. Together with equation~\eqref{eq:lnat},
we thus have that $\PopLoss{\nat}(\fstar)
= \PopLoss{\nat}(\fnat) \leq \PopLoss{\nat}(f)$ for all $f\in\Fspace$
by definition of $\fnat$.



If additionally the support of $P_z$ is equal to $\SubGorbit{z}$ and $\loss$ is injective,
the inequality~\eqref{eq:ineq} becomes a strict inequality for $\fnat\not \in \Fspace$
and hence we have $\PopLoss{\nat}(\fstar) < \PopLoss{\nat}(\fnat)$ which contradicts
the definition of $\fnat$ being the minimizer of the natural loss.

\section{Two-stage STN}
\label{sec:twostage}

Since STNs are known to be sensitive to hyperparameter settings and thus
 difficult to train 
end-to-end~\cite{Tai19}, we apply the following two-stage
procedure to simulate its functionality: (1) we first train a ResNet-32 as a localization regression
network (LocNet) to predict the attack perturbation separately
by learning from a training set, which contains perturbed images and uses the
transformations as the prediction targets; (2) at the same time we train a
ResNet-32 classifier with data augmentation, namely random translations and 
rotations; (3) during the test phase, the output of the LocNet is used by a
spatial transformer module that transforms the image before entering
the pretrained classifier. We refer to this two-stage STN as \twostage.

\paragraph{LocNet and Classifier} For the classifiers, we take the two models trained on CIFAR-10 and
SVHN using standard data augmentation and random rotations from our
previous experiments. Since we do not expect the regressors (or
LocNets) to be perfect in terms of prediction capability, there will
still be some transformation left after the regression stage.  Thus,
the classifiers should effectively see a smaller range of
transformations than without the inclusion of a LocNet and transformer module.
The training procedure used to train the classifiers is described in Section~\ref{sec:exp_details}.

\paragraph{Effect of rendering edges on LocNet} The LocNet is trained on zero padded
-- suffix $(c)$ -- as well as reflect padded inputs -- suffix $(r)$
-- for comparison.
 The former possibly yields an unfair advantage of this approach compared to other methods as the neural network can exploit the edges (induced through zero padding) to learn the transformation parameters.
Therefore, we also consider reflection padding to assess the effect of the different paddings on final performance.  
Nonetheless, zero padding is consistent with the augmentation
setting for the end-to-end trained networks and regularized methods
and was also the choice considered by~\cite{Engstrom17}.
For completeness we also show results when using reflection padding for
training LocNet although it lacks comparability with the other
methods since attacks should be reflection-padded as well.


\paragraph{Minimizing loss of information in the prediction transformation process} In the spatial transformer module we compare two variants of handling the
labels predicted by the LocNet. We can either back-transform the
transformed image with the negative predicted labels, which
will, under the assumption that the regressor successfully learnt
object orientations, turn back the image but potentially result in
extra padding space before we feed the images into the
classifier. Alternatively, we can subtract the predicted
transformation from the attack transformation, then use the remaining
transformation as the new ``attack transformation''.  The latter will
result in much smaller padding areas, if the LocNet is performing
well. From the experimental results we do see a big drop if we naively
transform images twice.  We denote the former method as ``naive'' and
latter as ``trick''.

\paragraph{Observed results}
For CIFAR-10, this two-stage classifier achieved relatively high grid
accuracies. However, the obtained accuracies are still lower than
expected, given that the LocNet is allowed to learn rotations with a
separately trained regressor on the transformed training set. For SVHN
we also see a gain compared to adversarial training without
regularizer. However, the performance still lags behind the accuracies
obtained by the regularizers. The results are summarized in
Table~\ref{tab:two-stage-eval}.


\begin{table}[htbp]
    \caption{Accuracies of two-stage \stn\, (\twostage) under different settings. Details are provided in Section~\ref{sec:twostage}. \label{tab:two-stage-eval}}
\begin{center}
\vskip 0.15in
 {\setlength{\tabcolsep}{4pt}
   \begin{tabular}{c c c c c }
Dataset & \twostage(c) trick & \twostage(r) trick & \twostage(c) naive & \twostage(r) naive\\
\hline
SVHN (nat) &94.92 & 95.51 & 94.92 & 95.51\\
 \qquad\quad (rob) &90.95 & 90.28 & 64.91 & 59.68\\
\hdashline
CIFAR10 (nat) &91.29 & 90.99 & 91.29 & 90.99\\
 \qquad\quad (rob) &83.05 & 84.31 & 44.88 & 42.84\\
\hline
    \end{tabular}}
  \end{center}
\end{table}

\section{More experimental results}
In this section we discuss additional experimental results that we collected and and analyzed.

\subsection{Stability to selection of regularization parameter $\lambda$}

\begin{figure*}[!htp]
\begin{center}
\begin{tabular}{ccccc}
\includegraphics[scale=0.44, keepaspectratio=true, trim={0 0 0 135}, clip]{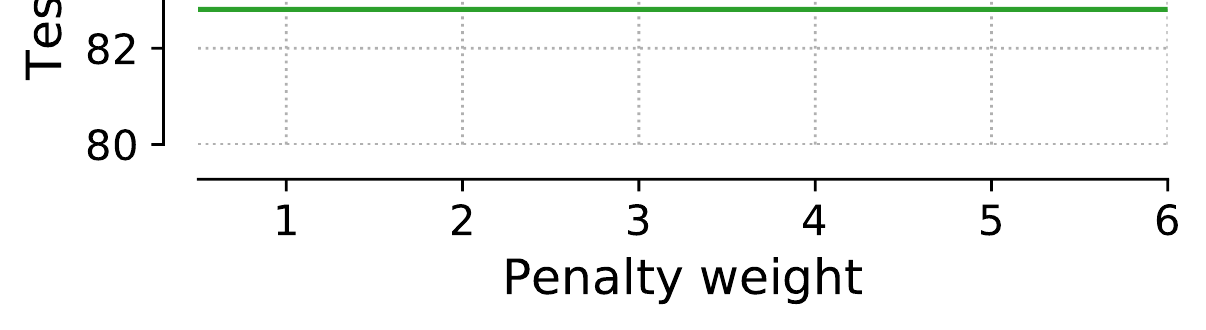} &
\includegraphics[scale=0.44, keepaspectratio=true, trim={0 0 0 135}, clip]{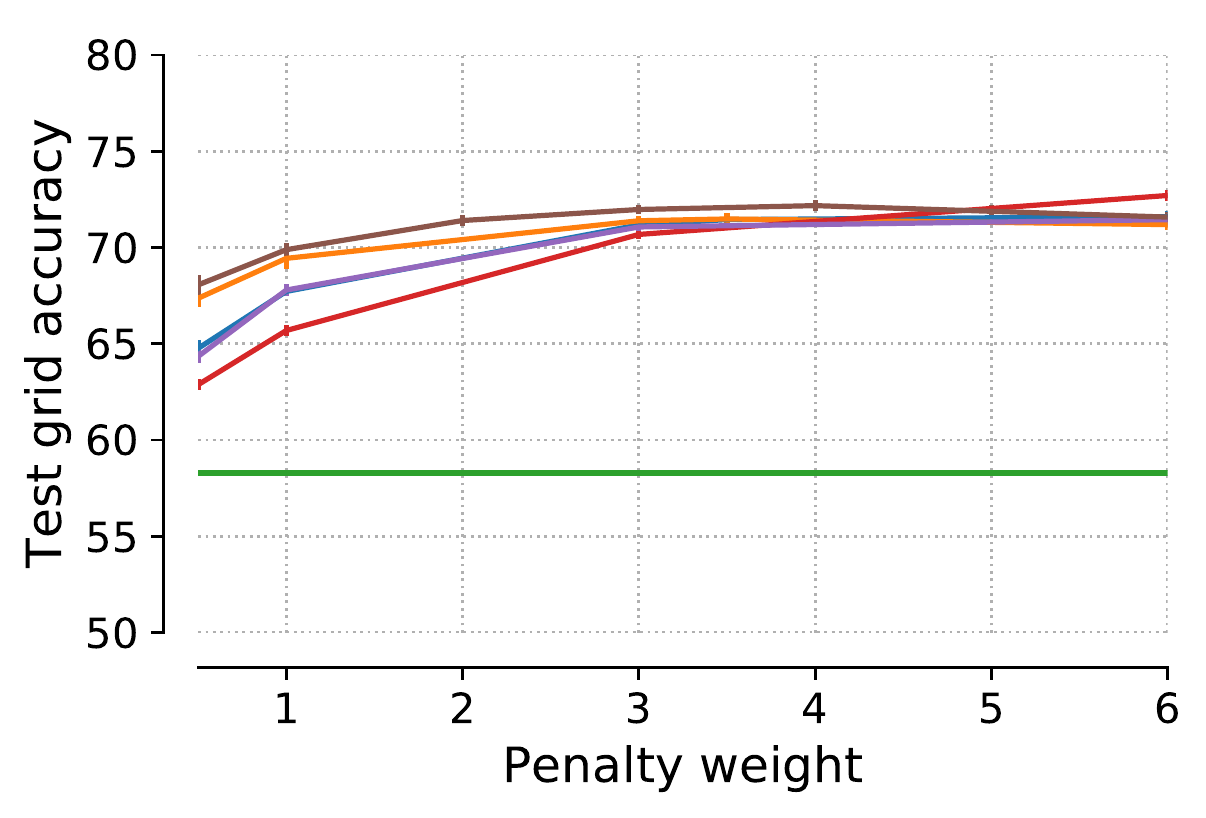} &
\includegraphics[scale=0.5, keepaspectratio=true, trim={0 0 0 0}, clip]{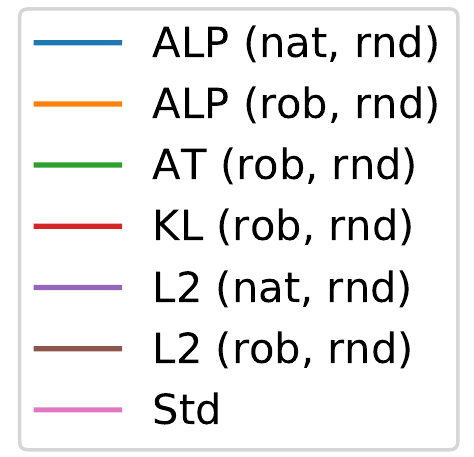} \\
SVHN & CIFAR-10 \\
%
%
\includegraphics[scale=0.40, keepaspectratio=true, trim={0 0 0 135}, clip]{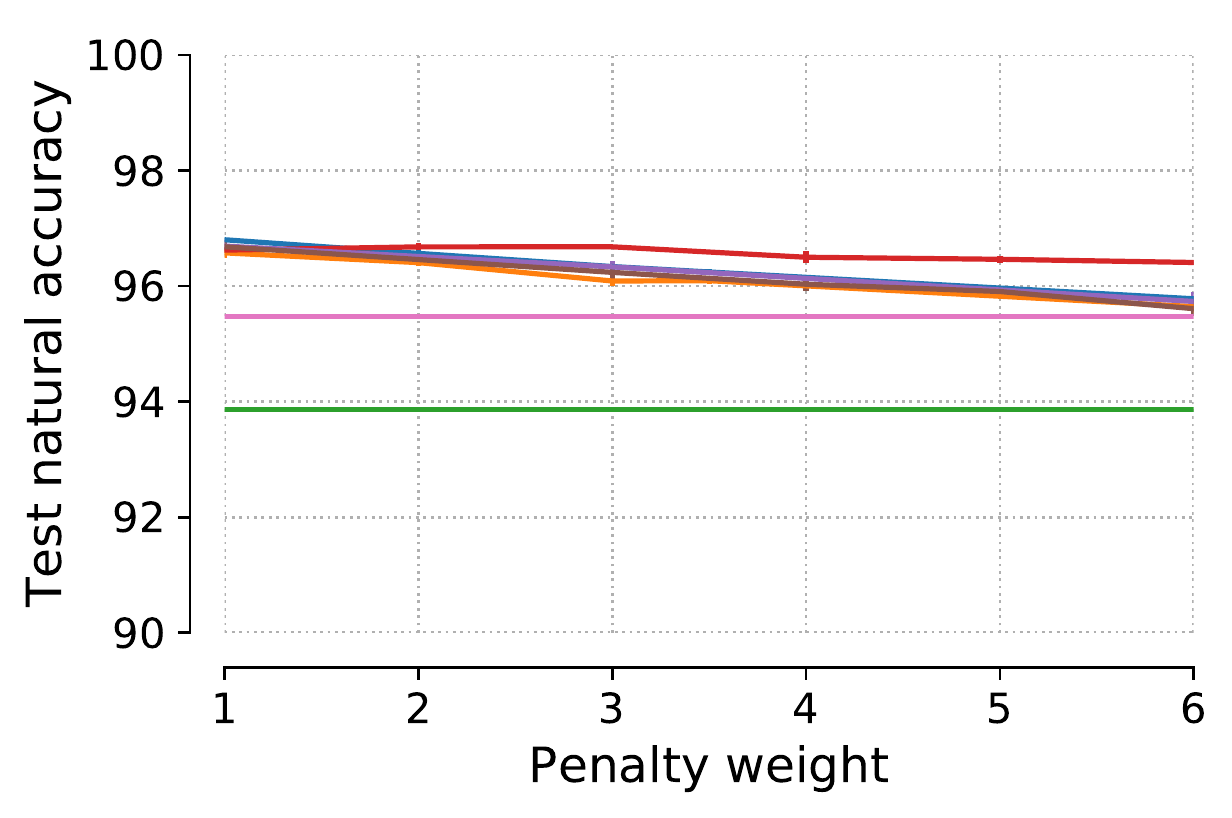} &
\includegraphics[scale=0.40, keepaspectratio=true, trim={0 0 0 135}, clip]{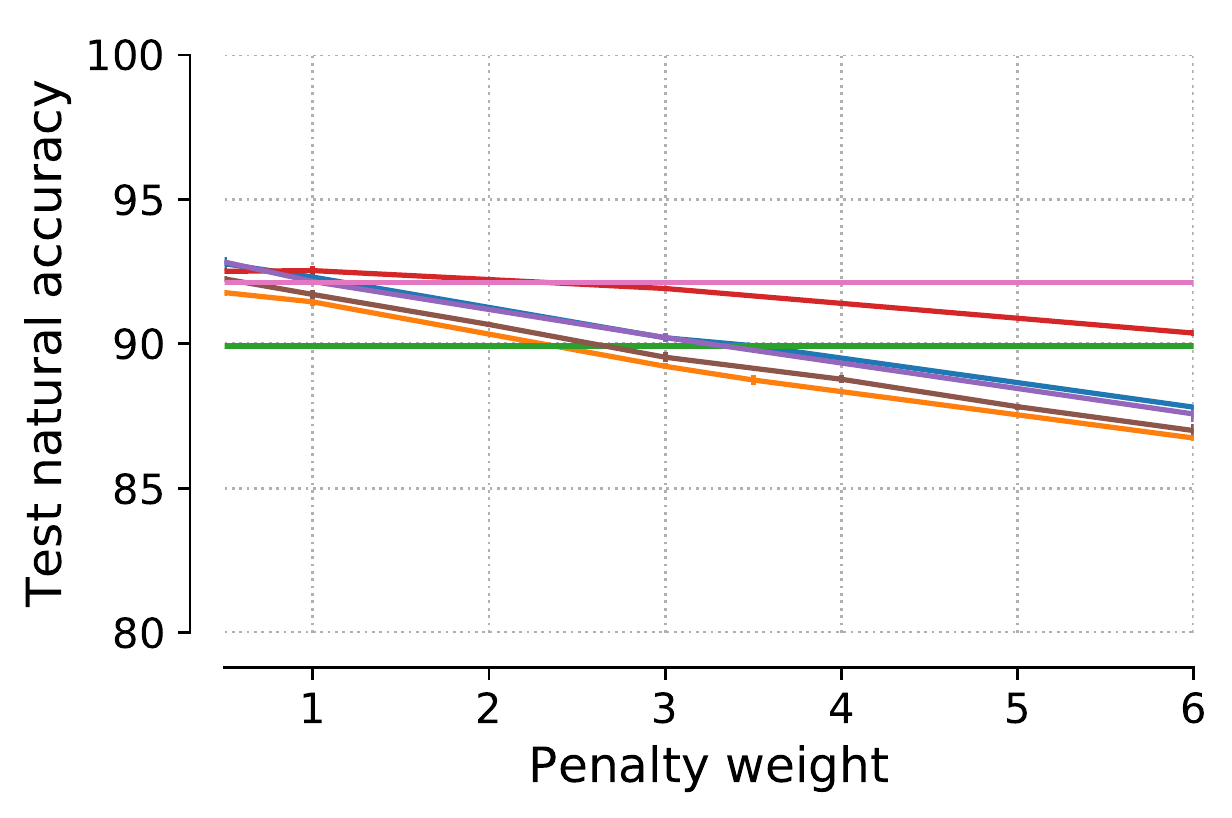} &
\\
SVHN & CIFAR-10 
\end{tabular}
\caption{Test grid accuracy (first row) and test natural accuracy (second row) as a function of the regularization parameter $\lambda$ for the SVHN (first column) and CIFAR-10 (second column) datasets and data augmentation (``\rnd''). The test grid accuracy is relatively robust in a large range of $\lambda$ values while natural test accuracy decreases with larger values of $\lambda$.}
\label{fig:grid_nat_overlambda_rnd}
\end{center}
\end{figure*}

\begin{figure*}[!htp]
\begin{center}
\begin{tabular}{ccccc}
\includegraphics[scale=0.44, keepaspectratio=true, trim={0 0 0 135}, clip]{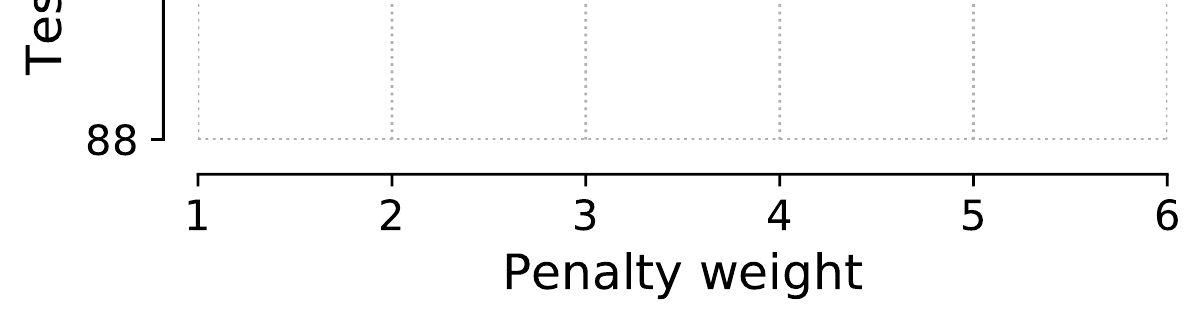} &
\includegraphics[scale=0.44, keepaspectratio=true, trim={0 0 0 135}, clip]{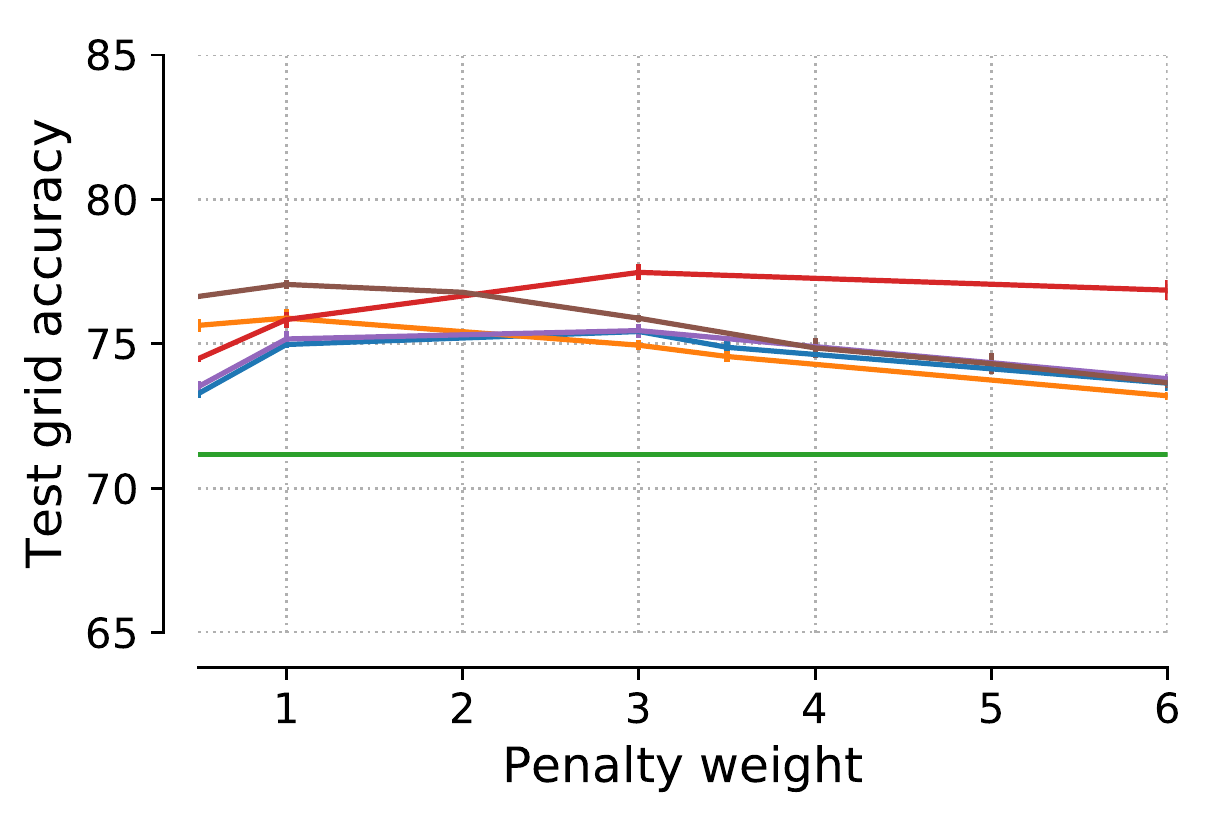} &
\includegraphics[scale=0.5, keepaspectratio=true, trim={0 0 0 0}, clip]{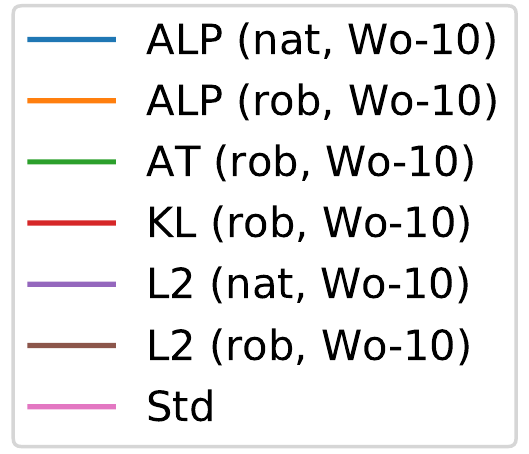}\\
SVHN & CIFAR-10 \\
%
%
\includegraphics[scale=0.44, keepaspectratio=true, trim={0 0 0 135}, clip]{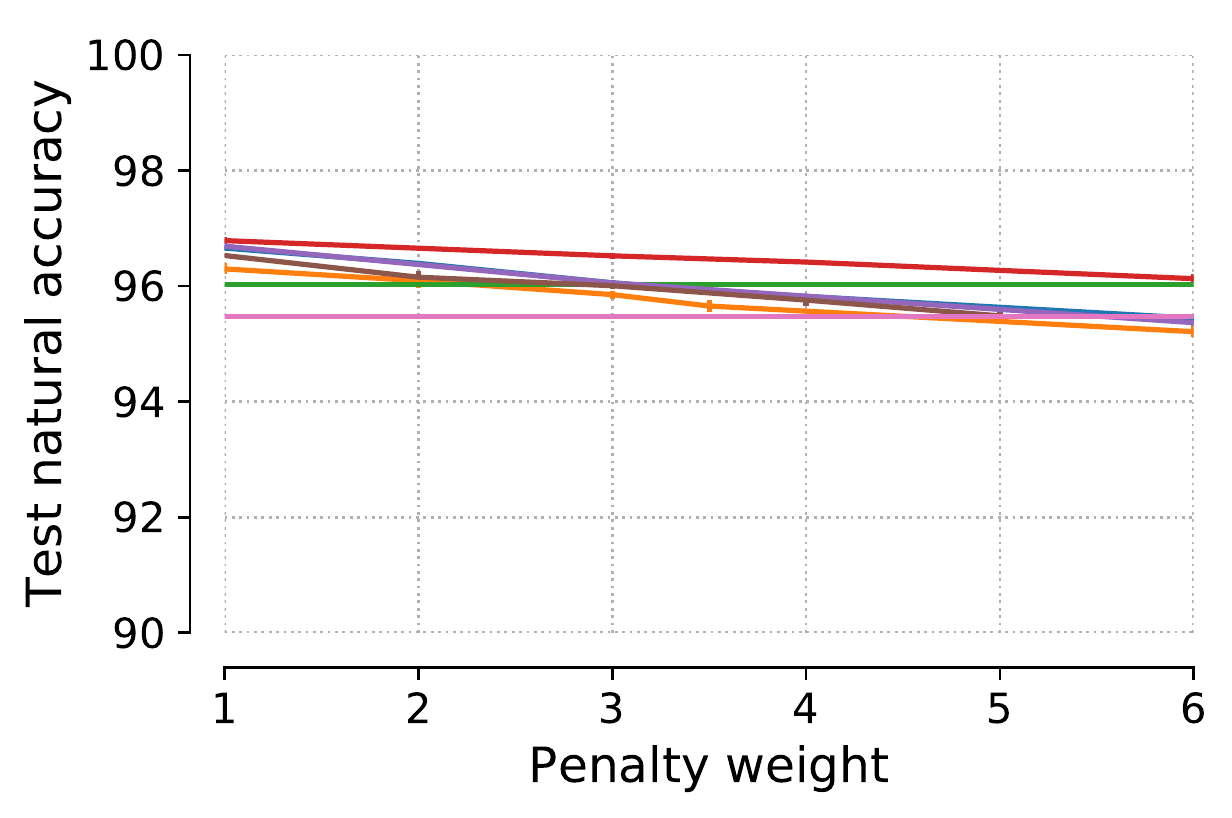} &
\includegraphics[scale=0.44, keepaspectratio=true, trim={0 0 0 135}, clip]{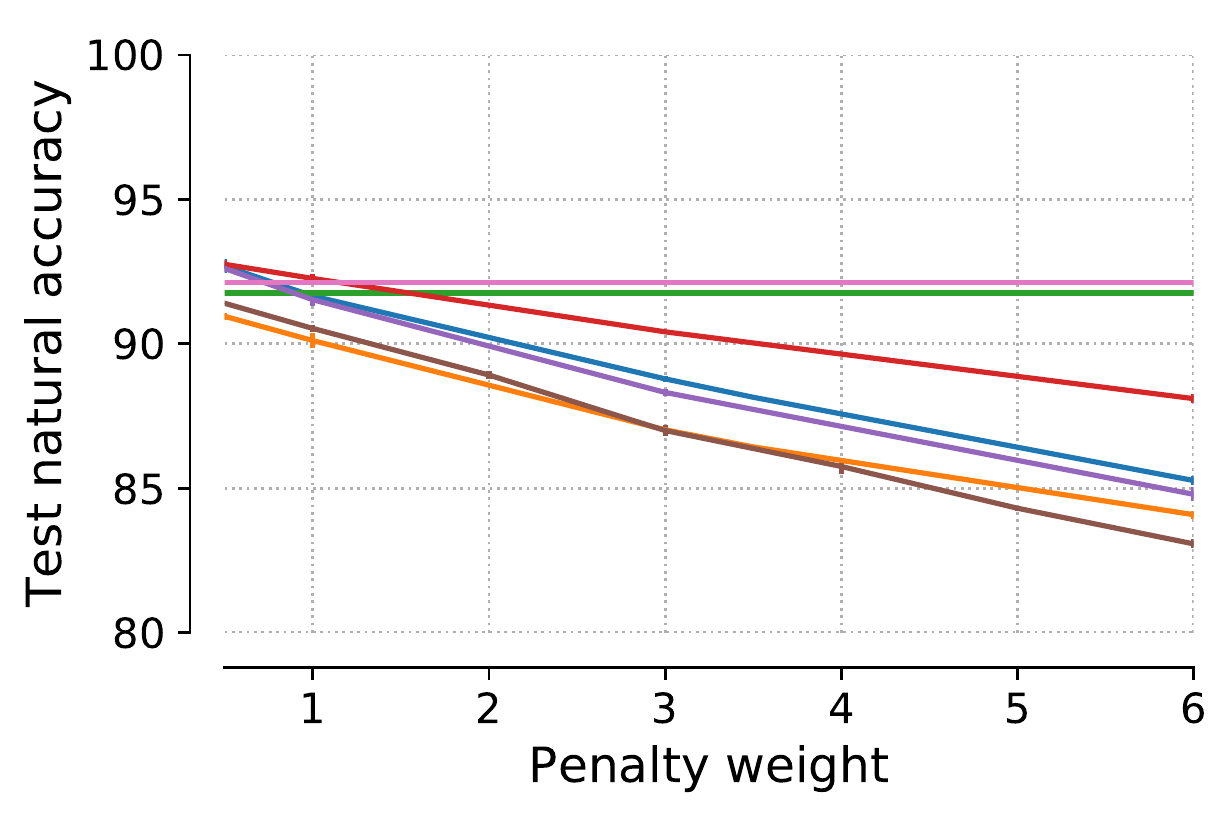} & \\
SVHN & CIFAR-10 
\end{tabular}
\caption{Test grid accuracy (first row) and test natural accuracy (second row) as a function of the regularization parameter $\lambda$ for the SVHN (first column) and CIFAR-10 (second column) datasets and $\wokabb$ defenses. The test grid accuracy is relatively robust in a large range of $\lambda$ values while natural test accuracy decreases with larger values of $\lambda$.}
\label{fig:grid_nat_overlambda_wo}
\end{center}
\end{figure*}

Figures~\ref{fig:grid_nat_overlambda_rnd} and~\ref{fig:grid_nat_overlambda_wo} show the test grid and test natural accuracy as a function of the regularization parameter $\lambda$. We observe that the regularization methods outperform unregularized methods in terms of grid accuracy in a large range of $\lambda$ values.

\subsection{Additional experimental results}
\label{sec:moreresults}

\begin{table}[H]
  \caption{Mean accuracies of models trained \textit{without} regularized adversarial training. Standard errors are shown in parentheses. \label{tab:compareadv1}}
\begin{center}
 {\setlength{\tabcolsep}{4pt}
   \begin{tabular}{c c c c c}
   & std & std* & AT(rob, Wo-10) & AT(mix, \woabb{10})\\
  \hline
  SVHN (nat) & 95.48 (0.15) & 93.97 (0.09) &96.03 (0.03) & 96.56 (0.07)\\
   \qquad\quad(rob) & 18.85 (1.27) & 82.60 (0.23) & 90.35 (0.27) & 88.83 (0.10)\\
  \hdashline
  CIFAR-10 (nat) & 92.11 (0.18) & 89.93 (0.18) & 91.76 (0.23) & 93.44 (0.19)\\
   \qquad\quad\quad(rob) & 9.52 (0.66) & 58.29 (0.60) & 71.17 (0.26) & 68.14 (0.48) \\
    \hdashline
  CIFAR-100 (nat) & 70.23 (0.18) & 66.62 (0.37) & 68.79 (0.34) & 73.03 (0.13)\\
   \qquad\quad\quad(rob) & 5.09 (0.25) & 28.53 (0.25) & 38.21 (0.10) & 35.93 (0.24)\\
  \hline

 \end{tabular}}
\end{center}
\end{table}
\vspace{-0.2in}

\paragraph{Mixed batch experiments} 
In addition to the results reported in the main text, in this section we also report results on more experiments that use the ``mixed batch'' setting, meaning that the gradient of the loss is taken with respect to both the adversarial and natural examples. This is common practice in the $\ell_p$ adversarial example literature \cite{Kannan18} and we denote this approach by ``mix''. As can be seen in Table~\ref{tab:compareadv1}, for adversarial training a mixed batch improves natural accuracy at the expense of test grid performance. For the regularization methods, we observe a much small, and not consistent, effect of the batch type as can be seen in Table~\ref{tab:compareadv3}. For example, comparing \alp(\rob, $\cdot$) vs.\ \alp(\mix, $\cdot$) shows that the performance differences are mostly not significant.

\begin{table}[H]
  \caption{Mean accuracies of models trained with various forms of
  regularized adversarial training, using the \tr\, regularization
  function. Standard errors are shown in
  parentheses. \label{tab:compareadv2}}
\begin{center}
\vskip 0.15in
 {\setlength{\tabcolsep}{3pt}
   \begin{tabular}{c c c c c c }
   & \makecell{\tr(\nat, \\rnd)} & \makecell{\tr(\nat,\\ Wo-10)} & \makecell{\tr(rob,\\ Wo-10)} & \makecell{\trcce(mix,\\ S-PGD)} & \makecell{KL(nat,\\ S-PGD)}  \\
  \hline
  SVHN (nat) &  96.16 (0.10) & 96.00 (0.02) & 96.13 (0.07) & 96.14 (0.04) & 96.54 (0.01) \\
   \qquad\quad (rob) &  90.69 (0.05) & 92.27 (0.09) & 92.71 (0.09) & 92.42 (0.03) &{\bf 92.62} (0.03) \\
  \hdashline
  CIFAR-10 (nat) &  89.33 (0.16) & 90.83 (0.18) & 90.41 (0.05) & 89.98 (0.21) & 89.82 (0.13) \\
   \qquad\quad\quad (rob) &  73.50 (0.19) & 77.34 (0.19) & 77.47 (0.28) & {\bf 78.93} (0.23) & 78.89 (0.07)\\
  \hline

 \end{tabular}}
\end{center}
\end{table}

\begin{table}[H]
  \caption{Mean accuracies of models trained with various forms of regularized adversarial training, using the \ltwo\, and \ALP\, regularization functions. Standard errors are shown in parentheses. \label{tab:compareadv3}}
\begin{center}
\vskip 0.15in
 {\setlength{\tabcolsep}{3pt}
   \begin{tabular}{c c c c c c c c}
   & \makecell{\ltwo(\nat, \\ \woabb{10})} & \makecell{\ltwo(\rob,\\ Wo-10)} & \makecell{\alp(mix,\\ Wo-10)}  & \makecell{\alp(\rob,\\ Wo-10)} & \makecell{\alp(mix,\\ Wo-20)} & \makecell{\alp(\rob,\\ S-PGD)} & \makecell{\alp(mix,\\ S-PGD)} \\
  \hline
  SVHN (nat) &  96.05 (0.04) &96.53 (0.03) & 96.41 (0.07) & 96.3 (0.09) & 96.39 (0.04) & 96.11 (0.08)   & 96.30 (0.09)\\
   \qquad\quad (rob) & 92.16 (0.05) &92.55 (0.08) & 92.17 (0.11) & 92.04 (0.19) & 92.48 (0.05) &92.32 (0.17)& 92.42 (0.20)\\
  \hdashline
  CIFAR-10 (nat) &  88.32 (0.13) & 90.53 (0.16) & 91.13 (0.13) & 90.11 (0.25) & 90.67 (0.12) & 89.85 (0.27) & 89.70 (0.10)\\
   \qquad\quad\quad (rob) &  75.46 (0.25) & 77.06 (0.16)& 75.89 (0.23)	& 75.90 (0.31) & 76.72 (0.21)	& 77.80 (0.17) & 77.72 (0.35)\\
   \hdashline
   CIFAR-100 (nat) &  - & - & 68.54 (0.27) & - & 68.04 (0.27) & 89.82 (0.13) & 68.44 (0.39)\\
   \qquad\quad\quad (rob) &  - & - & 49.30 (0.33) & - & 49.98 (0.31) & 78.89 (0.07) & 52.58 (0.20)\\
  \hline

 \end{tabular}}
\end{center}
\end{table}

\begin{table}[H]
  \caption{Mean standard and grid (\rob) accuracies of models trained with various forms of regularized adversarial training, using the \rnd\:(equivalent to Wo-1), Wo-10 and Wo-20 defense mechanisms for \tr\:(left) and \ALP\:(right). Standard errors are shown in parentheses. \label{tab:wok_diffk}}
  \begin{center}
\vskip 0.15in
 {\setlength{\tabcolsep}{3pt}
   \begin{tabular}{c c c c : c c c }
   & \makecell{\tr(\nat,\\ \woabb{1})} & \makecell{\tr(\nat, \\ \woabb{10})} & \makecell{\tr(\nat,\\ \woabb{20})}  & \makecell{\alp(\rob,\\ \woabb{1})} & \makecell{\alp(\rob,\\ \woabb{10})} & \makecell{\alp (\rob,\\ \woabb{20})} \\
  \hline
  CIFAR-10 (nat) &  89.34 (0.16) & 90.83 (0.18) & 89.33 (0.22) & 89.47 (0.04) & 90.11 (0.25) & 90.62 (0.07)\\
   \qquad\quad (rob) & 73.40 (0.19) & 77.34 (0.19) & 77.52 (0.16) & 73.22 (0.14) & 75.90 (0.31) & 76.78 (0.15)\\
  \hline
 \end{tabular}}
\end{center}
\end{table}

\begin{table}[H]
  \caption{Mean accuracies of models trained with various forms of augmented training, i.e.\ unregularized and regularized data augmentation. Standard errors are shown in parentheses. \label{tab:compareadv4}}
\begin{center}
\vskip 0.15in
 {\setlength{\tabcolsep}{3pt}
   \begin{tabular}{c c c c c c c}
   &  std* &  \ltwo(nat, rnd) & KL(nat, rnd) & ALP(rob, rnd) & KL(rob, rnd) & ALP(mix, rnd) \\
  \hline
  SVHN (nat) &  93.97 (0.09)&96.34 (0.08)&96.16 (0.10)&96.09 (0.06)&96.23 (0.08)&96.19 (0.07)   \\
   \qquad\quad (rob) & 82.60 (0.23)&90.51 (0.15)&90.69 (0.05)&90.48 (0.16)&90.92 (0.17)&90.48 (0.15)\\
  \hdashline
  CIFAR-10 (nat) & 89.93 (0.18)&87.80 (0.11)&89.34 (0.16)&88.75 (0.18)&89.47 (0.04)&89.43 (0.28) \\
   \qquad\quad\quad (rob) &  58.29 (0.60)&71.60 (0.27)&73.50 (0.19)&71.49 (0.30)&73.22 (0.14)&71.97 (0.11)\\
  \hline
 \end{tabular}}
\end{center}
\end{table}

\vspace{-0.2in}
\begin{table}[H]
  \caption{Mean accuracies of models trained with various forms of regularized adversarial training. Standard errors are shown in parentheses. \label{tab:compareadv5}}
\begin{center}
\vskip 0.15in
 {\setlength{\tabcolsep}{4pt}
   \begin{tabular}{c c c c c c}
   & \ALP(\nat, \woabb{10})& \ltwo(\nat, \woabb{10})&	\trcce(nat, \woabb{10}) &	\tr(\nat, \woabb{10}) \\
  \hline
  SVHN (nat) &  96.39 (0.03)&	96.05 (0.04) &96.18 (0.06)&	96.00 (0.02)   \\
   \qquad\quad (rob) & 91.98 (0.13)	&92.16 (0.05)	&91.99 (0.12)&	92.27 (0.09)\\
  \hdashline
  CIFAR10 (nat) & 88.78 (0.11) &	88.32 (0.13)&	89.61 (0.09)&	90.83 (0.18) \\
   \qquad\quad\quad (rob) &  75.43 (0.13)&	75.46 (0.25)&	76.15 (0.23)&	77.34 (0.19)\\
  \hline
 \end{tabular}}
\end{center}
\end{table}

\paragraph{Weakness of first order attack.} Table~\ref{tab:fo_supp} shows the accuracies of various models trained with $\fo$ defenses and evaluated against the $\fo$ and the grid search attack on all datasets. We observe that the $\fo$ attack constitutes are very weak attack since the associated accuracies are much larger than for the grid search attack. In other words, the $\fo$ attack only yields a very loose upper bound on the adversarial accuracy. This stands in stark contrast to $\ell_\infty$ attacks and has first been noted and discussed in \cite{Engstrom17}. Interestingly, using the first order method as a {\it defense} mechanism proves to be very effective in terms of grid accuracy. When used in combination with a regularizer this defense yields the largest overall accuracies as shown and discussed in Section~\ref{sec:results}. Recall that due to computational reasons grid search cannot be used as a defense mechanism. Therefore, the strongest computationally feasible defense does not use the same mechanism as the strongest attack in our setting.
\begin{table}[htbp]
	\begin{center}
  \vskip 0.15in
   {\setlength{\tabcolsep}{4pt}
	 \begin{tabular}{c c c c c }
	 & $\AT(\mix,{\fo})$& $\AT(\rob,{\fo})$&	$\ALP(\mix,{\fo})$ &\\
	\hline
	\quad SVHN (nat) &  96.27 (0.00)&	96.06 (0.10) &96.30 (0.09)\\ 
	 \qquad\qquad (grid) & 84.81 (0.01)	&87.29 (0.09)	&{\bf 92.42 (0.20)}\\
	 \qquad\quad ($\fo$) &  95.26 (0.04)&	95.46 (0.10) &95.92 (0.13)   \\
	 \hdashline
	CIFAR-10 (nat) & 92.19 (0.23)&	91.83 (0.19)&	89.70 (0.10) \\ 
	 \qquad\quad\quad (grid) &  64.26 (0.25)&	69.74 (0.27) &	{\bf 77.72 (0.35)}\\
	 \qquad\quad ($\fo$) &  88.84 (0.27)&	89.87 (0.10) &88.15 (0.21)  \\
	 \hdashline
	CIFAR-100 (nat) & 71.11 (0.37)&	68.87 (0.19)&	68.44 (0.39) \\ 
	 \qquad\quad\quad (grid) &  33.40 (0.21)&	37.87 (0.12)&	{\bf 52.58 (0.20)}\\
	 \qquad\quad ($\fo$) &  65.01 (0.32)&	65.56 (0.12) &66.04 (0.40)  \\
	 \hline
   \end{tabular}}
  \end{center}
      \caption{Mean accuracies of different models trained with $\fo$ defenses and evaluated on the natural test set, against the $\fo$ attack and against the grid search attack on the SVHN, CIFAR-10 and CIFAR-100 datasets. While the test accuracy for the $\fo$ attack is only slightly lower than the natural accuracy in most cases, the grid accuracy is significantly smaller.  \label{tab:fo_supp}}
  \end{table}

\newpage
\paragraph{Stronger grid search attack}
To evaluate how much grid accuracy changes with a finer discretization of the perturbation set $\Pertset$, we compare the default grid to a finer one for a subset of experiments, summarized in Table~\ref{tab:finer_grid}. Specifically, ``(grid-775)" shows the test grid accuracy using the default grid containing 5 values per translation direction and 31 values for rotation, yielding a total of 775 transformed examples that are evaluated for each $X_i$. ``(grid-7500)'' shows the test grid accuracy on a much finer grid with 10 values per translation direction and 75 values for rotation, resulting 7500 transformed examples. We observe that the test grid accuracy only decreases slightly for the finer grid and the reduction in accuracy is smaller for $\ALP$ than for $\AT$. Due to computational reasons we use the grid containing 775 values for all other experiments.

\begin{table}[H]

	\begin{center}
  \vskip 0.15in
   {\setlength{\tabcolsep}{4pt}
	 \begin{tabular}{c c c c c }
	 & $\AT(\mix,{\woabb{10}})$& $\AT(\rob,{\woabb{10}})$&	$\ALP(\mix,\woabb{10})$ &\\
	\hline
	\quad SVHN (grid-775) &  88.83 (0.10)&	89.75 (0.17) &92.17 (0.11)\\
	 \qquad\qquad (grid-7500) & 88.02 (0.12)	&89.29 (0.15)	&91.79 (0.12)\\
	 	 \hdashline
	CIFAR-10 (grid-775) & 68.14 (0.48)&	70.35 (0.16)&	75.89 (0.23) \\
	 \qquad\quad\quad (grid-7500) &  65.69 (0.28)&	68.28 (0.16)&	74.58 (0.16)\\
	 	 \hdashline
	CIFAR-100 (grid-775) & 35.93 (0.24)&	38.21 (0.10)&	49.30 (0.33) \\
	 \qquad\quad\quad (grid-7500) &  33.62 (0.23)&	36.04 (0.21)&	47.95 (0.23)\\
		 \hline
   \end{tabular}}
  \end{center}
  	\caption{Mean accuracies for different models evaluated against two different grid search attacks. grid-775 represents test grid accuracy using the default grid with 775 transformed examples, grid-7500 shows test grid accuracy on a much finer grid with 7500 transformed examples. Test grid accuracy only decreases slightly for the finer grid and the reduction in accuracy is smaller for $\ALP$ than for $\AT$.  \label{tab:finer_grid}}
  \end{table}

\clearpage
\newpage
\subsection{Regularization effect on range of incorrect angles}
\vspace{-0.2in}
\begin{figure*}[ht]
\begin{center}
\begin{tabular}{cccc}
\hspace{-.2cm}
\includegraphics[scale=0.45, keepaspectratio=true, trim={0 0 0 0}, clip]{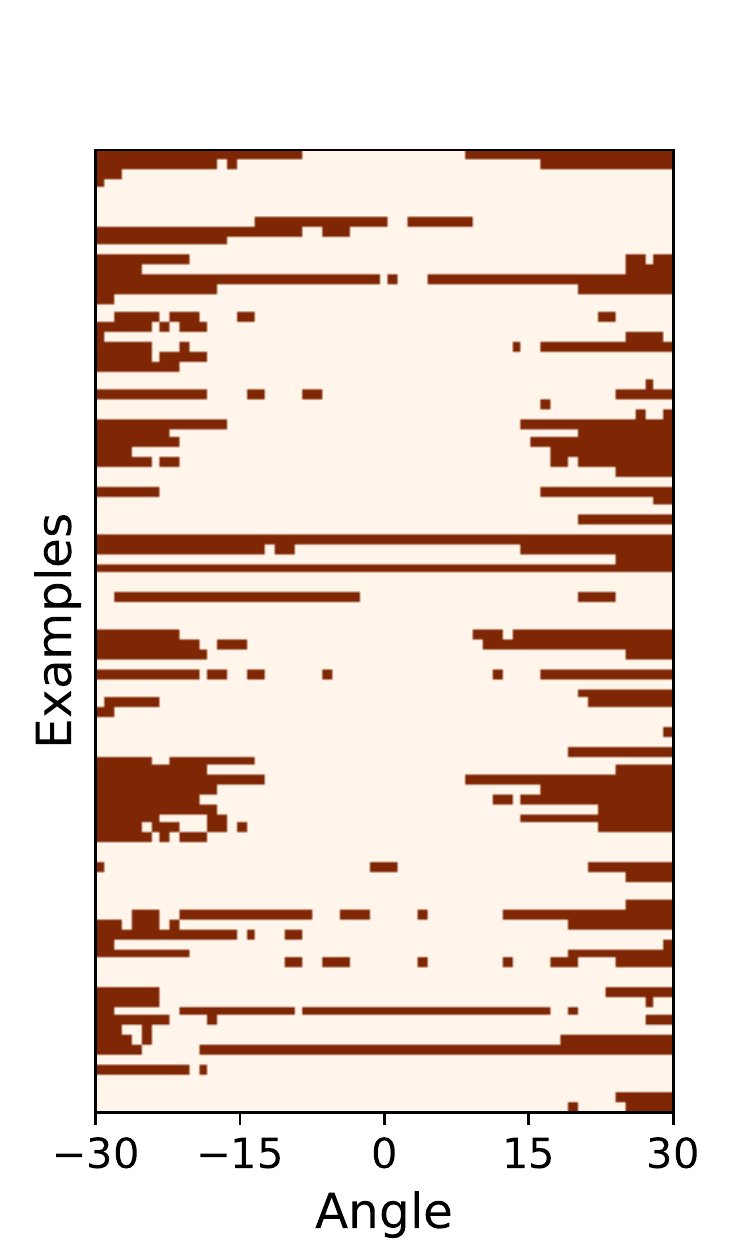} &
\hspace{-0cm}
\includegraphics[scale=0.45, keepaspectratio=true, trim={0 0 0 0}, clip]{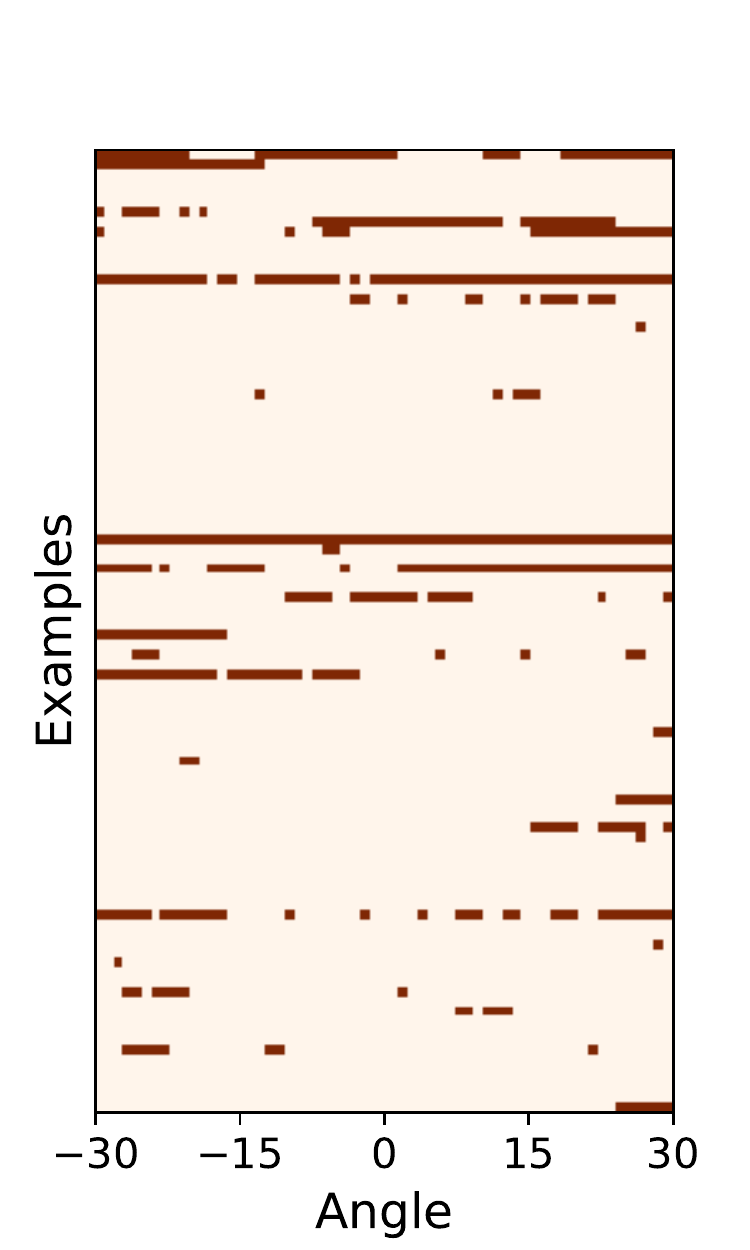} &
\hspace{-.4cm}
\includegraphics[scale=0.45, keepaspectratio=true, trim={0 0 0 0}, clip]{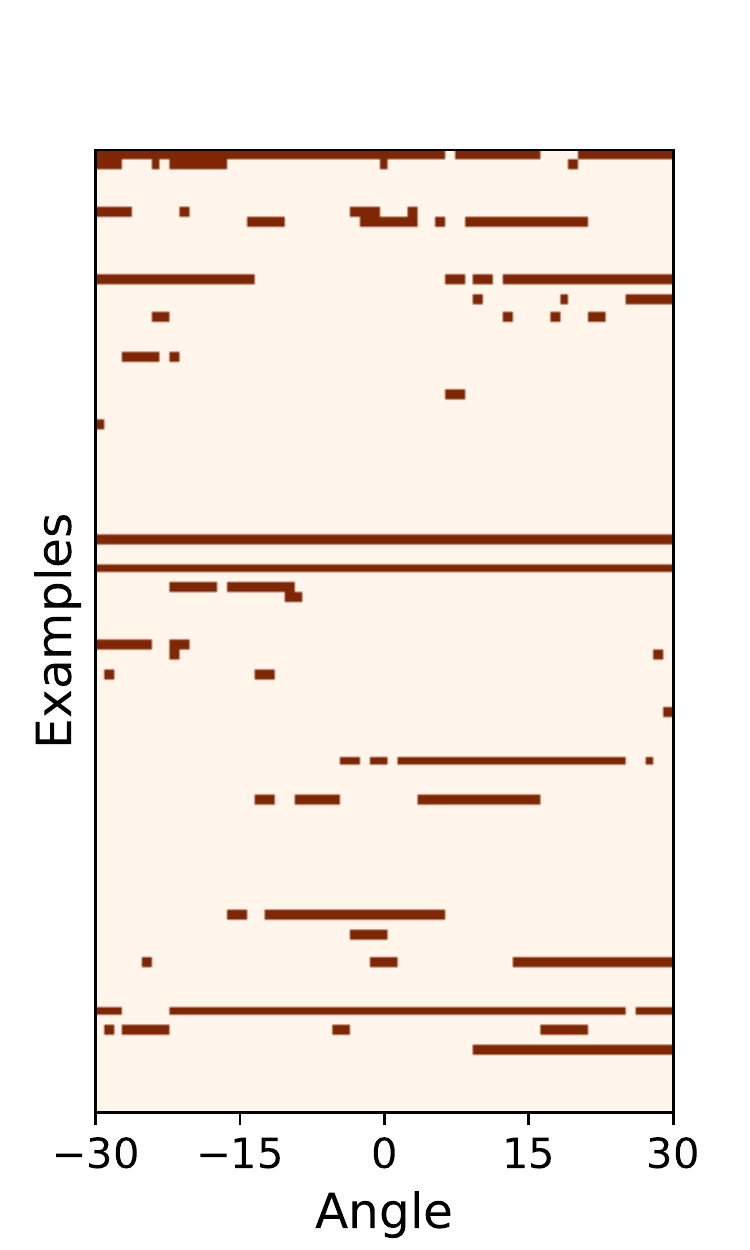} &
\hspace{-.5cm}
\includegraphics[scale=0.45, keepaspectratio=true, trim={0 0 0 0}, clip]{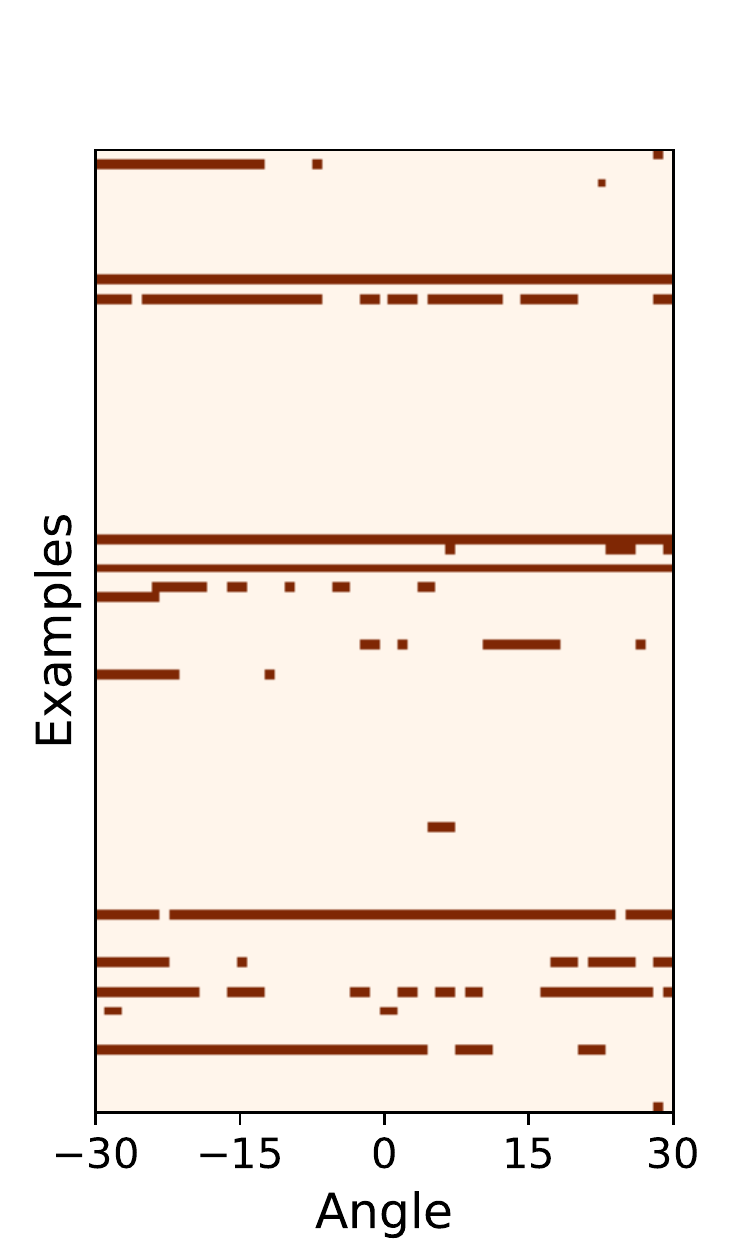} \\
{\small Standard} & {\small$\AT(\mix,\rnd)$} & {\small$\AT(\rob,\rnd)$} &  {\small$\ALP(\mix,\rnd)$}
\end{tabular}
%
%
\begin{tabular}{cccc}
\hspace{3.4cm}
&
\hspace{-0cm}
\includegraphics[scale=0.45, keepaspectratio=true, trim={0 0 0 0}, clip]{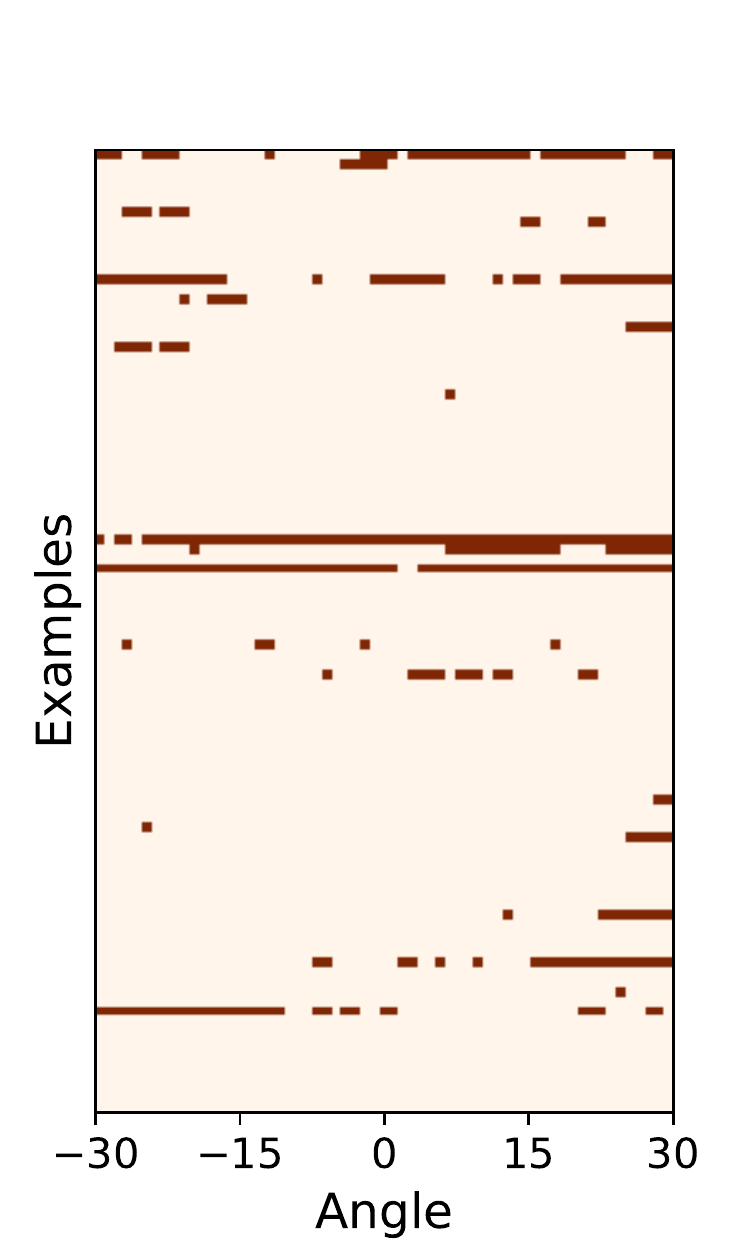} &
\hspace{-.4cm}
\includegraphics[scale=0.45, keepaspectratio=true, trim={0 0 0 0}, clip]{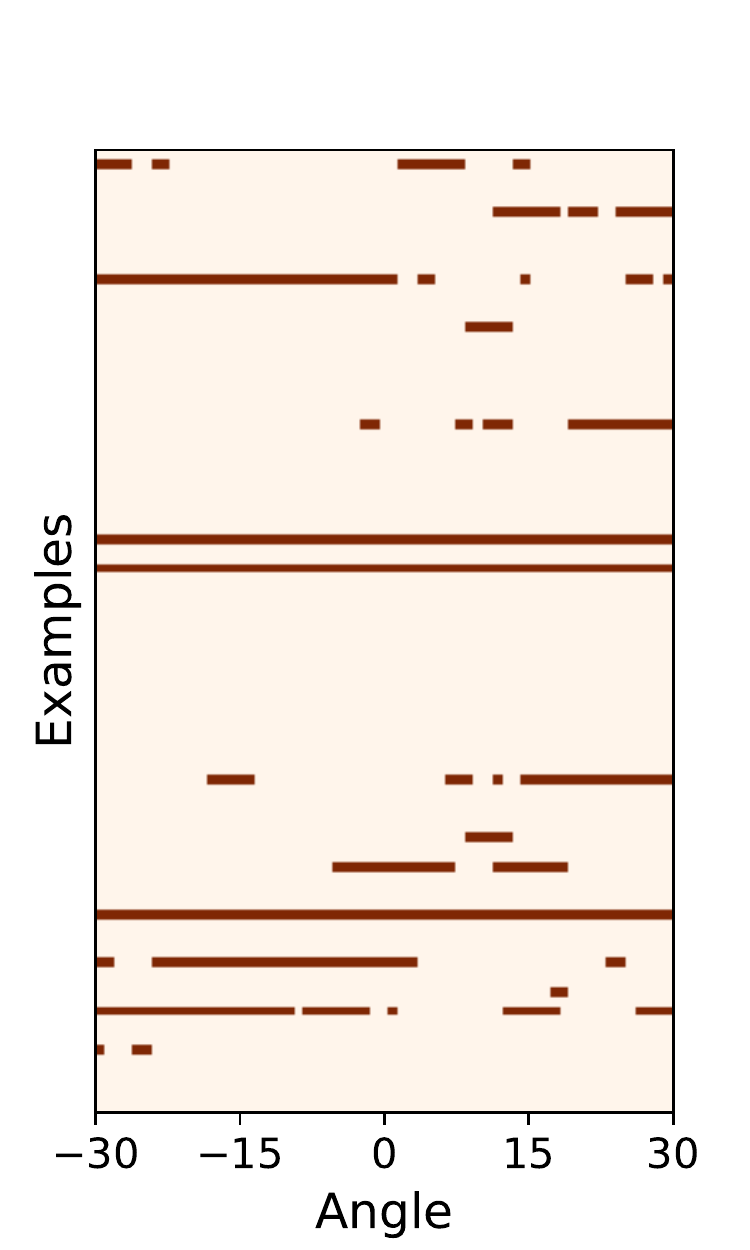} &
\hspace{-.5cm}
\includegraphics[scale=0.45, keepaspectratio=true, trim={0 0 0 0}, clip]{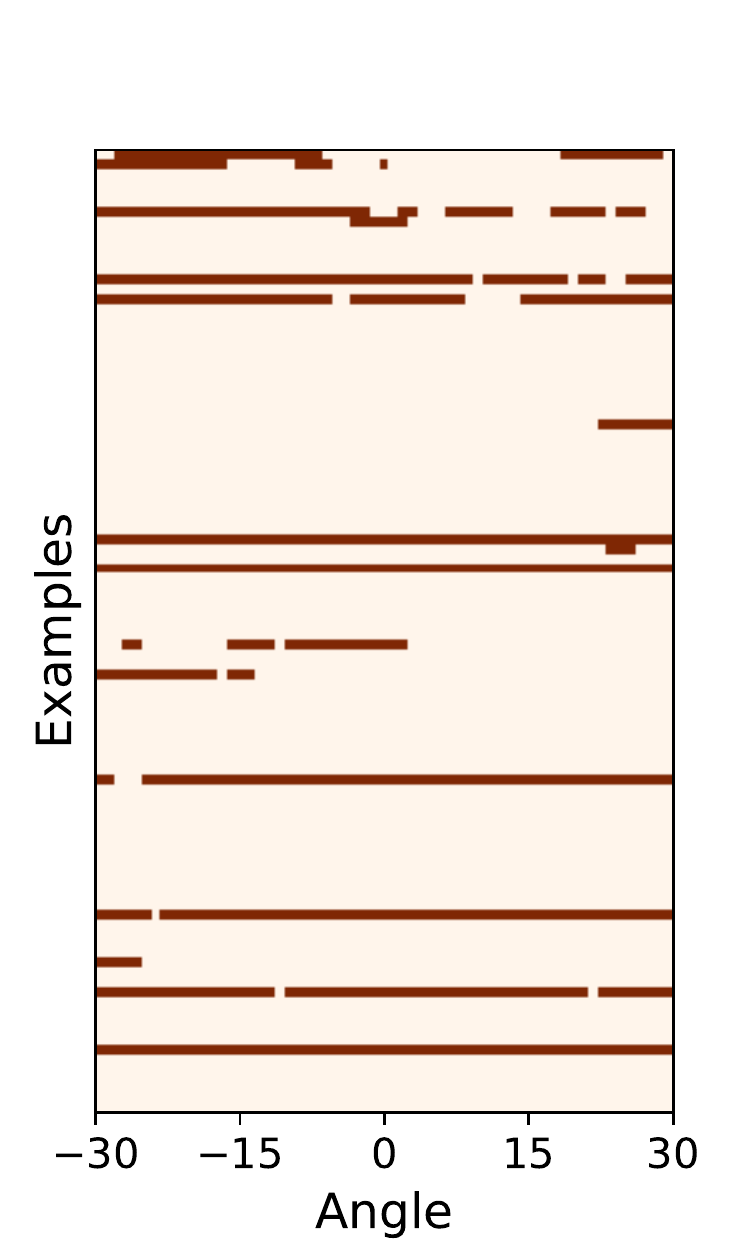} \\
{  }& {\small$\AT(\mix,{\woabb{10}})$} & {\small$\AT(\rob,{\woabb{10}})$} & {\small$\ALP(\mix,{\woabb{10}})$}
\end{tabular}
\caption{For 100 randomly chosen examples from the CIFAR-10 dataset, we show which rotations lead to a misclassification by various models. Each row corresponds to one example and each column to one angle in the interval $[-30^\circ, 30^\circ]$. A dark red square indicates that the corresponding example was misclassified after being rotated by the corresponding angle. The visualization for $\AT(\mix,\rnd)$ is more fragmented than for $\AT(\rob,\rnd)$ and $\ALP(\mix,\rnd)$ and the visualization for $\AT(\mix,{\woabb{10}})$ is more fragmented than for $\AT(\rob,{\woabb{10}})$ and $\ALP(\mix,{\woabb{10}})$.}
\label{fig:angles_k10}
\end{center}
\end{figure*}

\end{document}